\pdfoutput=1

\documentclass[11pt]{article}

\usepackage[final]{acl}

\usepackage{times}
\usepackage{latexsym}
\usepackage{float}
\usepackage{hyperref}
\usepackage{subfigure}

\usepackage{enumitem} 

\usepackage[most]{tcolorbox}  
\usepackage{xcolor}           

\usepackage{booktabs}
\usepackage{siunitx}

\usepackage[T1]{fontenc}

\usepackage[utf8]{inputenc}

\usepackage{microtype}

\usepackage{inconsolata}

\usepackage{graphicx}
\usepackage{xspace}
\usepackage{multirow}
\usepackage{longtable}

\newcommand{\benchmark}{\textit{CulturalPersonas}\xspace}

\title{Can LLMs Express Personality Across Cultures? \break Introducing \benchmark for Evaluating Trait Alignment}

\author{
Priyanka Dey \quad Yugal Khanter \quad Aayush Bothra \\ \textbf{Jieyu Zhao} \quad \textbf{Emilio Ferrara} \\
Department of Computer Science, University of Southern California \\
\texttt{deyp@usc.edu}
}

\begin{document}
\maketitle
\begin{abstract}

As LLMs become central to interactive applications, ranging from tutoring to mental health, the ability to express personality in culturally appropriate ways is increasingly important. While recent works have explored personality evaluation of LLMs, they largely overlook the interplay between culture and personality. To address this, we introduce \benchmark, the first large-scale benchmark with human validation for evaluating LLMs’ personality expression in culturally grounded, behaviorally rich contexts. Our dataset spans 3,000 scenario-based questions across six diverse countries, designed to elicit personality through everyday scenarios rooted in local values. We evaluate how closely three models' personality distributions align to real human populations through two evaluation settings: multiple-choice and open-ended response formats. Our results show-- \benchmark improves alignment with country-specific human personality distributions (over a 20\% reduction in Wasserstein distance across models and countries) and elicits more expressive, culturally coherent outputs compared to existing benchmarks. \benchmark surfaces meaningful modulate trait outputs in response to culturally grounded prompts, offering new directions for aligning LLMs to global norms of behavior. By bridging personality expression and cultural nuance, we envision that \benchmark will pave the way for more socially intelligent and globally adaptive LLMs. Datasets and code are available at: \href{https://github.com/limenlp/CulturalPersonas}{https://github.com/limenlp/CulturalPersonas}.

\end{abstract}

\section{Introduction}
\label{sec:intro}

From customer support agents to personal tutors, large language models (LLMs) are central to interactive technologies that engage users daily. These systems are increasingly expected to display not just fluency or task competence, but \textit{personality}: the ability to convey a psychological profile that users perceive as trustworthy and relatable.

In psychology and NLP, the Big Five (\texttt{OCEAN}) framework is widely used to model personality through five traits—\textbf{O}penness, \textbf{C}onscientiousness, \textbf{E}xtraversion, \textbf{A}greeableness, and \textbf{N}euroticism—that represent consistent patterns of thought, behavior, and emotion. Recent work evaluates LLMs using psychometric tools like the Big Five Inventory (BFI)~\cite{john1999bigfive} and the International Personality Item Pool (IPIP)~\cite{goldberg1999ipip}, via self-reflective statements (e.g., \textit{“I talk a lot”}, \textit{“I make people feel at ease”})~\cite{jiang-etal-2024-personallm, bodroza2024personality, salecha2024bias}. Moving beyond static questionnaires, newer benchmarks such as \textit{TRAIT}~\cite{lee-etal-2025-llms} and \textit{Big5Chat}~\cite{li2024big5chat} embed personality traits into dialogue or situational prompts, enabling more behaviorally grounded evaluations.

While these tools offer a useful starting point, they fall short of capturing how personality is expressed through behavior in real-world contexts. Moreover, they assume that personality is culturally universal despite decades of work in cultural psychology, which shows that trait expression is deeply shaped by local norms, values, and communicative styles~\cite{hofstede2001culture, han2023social, liu2024culturally}. For example, in Western cultures such as the United States, \textit{assertiveness} may reflect high openness. However, in nations such as Japan, this same quality may be seen as disruptive; instead, traits such as harmony and indirectness are stronger and more preferred indicators of \textit{Openness}. Evaluating LLMs without accounting for these distinctions risks reinforcing Western behavioral norms and misrepresenting how personality is interpreted across cultures.

To address this gap, we introduce \textbf{\benchmark}—the first large-scale benchmark for evaluating personality expression in LLMs through culturally grounded, behavior-rich scenarios. \benchmark comprises 3,000 situational prompts spanning six culturally diverse countries: Brazil, India, Japan, Saudi Arabia, South Africa, and the United States. Each prompt targets one of the Big Five traits by embedding it in realistic, locally situated social contexts that reflect cultural values—i.e., shared behavioral norms and expectations that shape how individuals act within a society. Scenarios are generated via a retrieval-augmented generation (RAG) pipeline \cite{lewis2020retrieval} seeded with academic cultural texts and subsequently validated by native annotators for cultural fidelity (see Figure~\ref{fig:dataset_creation_new} and §\ref{sec:human_val}).

We use \benchmark to measure the personality profiles of three models— GPT-4o-mini, Llama-3-8B, and Qwen2-7B— across two evaluation formats: multiple-choice selection and open-ended generation and compare alignment to real human populations from diverse cultural backgrounds. Our contributions are fourfold:

\begin{itemize}[leftmargin=10pt,noitemsep]

\item \textbf{Conceptual Reframing of Personality Evaluation:} Drawing from cultural psychology, we take a first step toward assessing personality expression in LLMs through a cultural lens by examining how well models represent the personalities associated with different national profiles. To the best of our knowledge, cross-cultural personality evaluation in LLMs remains underexplored, and our work highlights the need for more culturally grounded, behavior-focused evaluators (§\ref{sec:background}).

\item \textbf{Benchmark for Cultural Personality Expression:} We present \benchmark, the first benchmark to jointly evaluate personality traits and cultural context in LLMs (§\ref{sec:cp}). By embedding the Big Five traits in realistic, culturally grounded scenarios, \benchmark moves beyond generalized prompts to support more culturally relevant scenarios.

\item \textbf{Dual-Format Evaluation Framework:} We propose a two-part evaluation: multiple-choice selection (MCS) and open-ended generation (OEG)—that captures both trait alignment with real user populations and expressive diversity. This framework enables both distributional comparison with human norms and stylistic analysis of personality expression in various contexts (§\ref{sec:experiments}).

\item \textbf{Empirical Findings on Model Behavior:} Our results show that \benchmark improves alignment with human personality distributions over existing personality evaluators, e.g., \textit{IPIP} and \textit{TRAIT}, elicits more expressive and lexically diverse outputs, and surfaces model-specific differences in cultural adaptability (§\ref{sec:results}).

\end{itemize}

\section{Background}
\label{sec:background}

\subsection{Personality Evaluation Methods}

Personality can be defined through various psychological frameworks, including the Big Five \cite{john1999bigfive}, Myers-Briggs Type Indicator (MBTI) \cite{myers1987mbti}, and the Dark Triad \cite{paulhus2002darktriad}. We focus on the Big Five model—introduced in §\ref{sec:intro}—which is widely used for modeling trait-level dispositions. 

Traits are typically rated on a 1–5 Likert scale using standardized instruments like the Big Five Inventory (\textit{BFI}) \cite{fossati2011big}, \textit{IPIP-120}, and \textit{IPIP-300} \cite{goldberg1999ipip}, which use statement-based items such as \textit{“I talk a lot”} or \textit{“I have a lot of energy”}. \textit{BFI} includes 44 items, \textit{IPIP-120} has 120, and \textit{IPIP-300} consists of 300 statements. 

\begin{table}[!t]
\centering
\scriptsize
\setlength{\tabcolsep}{5.1pt}
\begin{tabular}{lccccc}
\toprule
\textbf{Country Pair} & \textbf{O} & \textbf{C} & \textbf{E} & \textbf{A} & \textbf{N} \\
\midrule
USA–Brazil        & 0.31$^{*}$ & 0.85$^{**}$ & 0.42$^{*}$ & 0.53$^{**}$ & 0.46$^{**}$ \\
Brazil–Saudi Arabia & 0.61$^{**}$ & 0.37$^{*}$ & 0.64$^{**}$ & 0.39 & 0.38$^{*}$ \\
South Africa–Japan & 0.61$^{**}$ & 0.28$^{*}$ & 0.38$^{*}$ & 0.58$^{**}$ & 0.29$^{*}$ \\
India–Saudi Arabia & 0.38$^{*}$ & 0.39$^{*}$ & 0.45$^{*}$ & 0.57$^{**}$ & 0.48$^{**}$ \\
\bottomrule
\end{tabular}
\caption{Pairwise Wasserstein distances ($W$) between selected country pairs for each \texttt{OCEAN} trait. Statistical significance is measured via 2-sample KS test (*p $<$ 0.05, **p $<$ 0.01). Larger $W$ indicates higher cross-cultural differences.} 
\label{tab:distances}
\end{table}

\begin{figure*}[!t]
\centering
\includegraphics[width=0.95\textwidth]{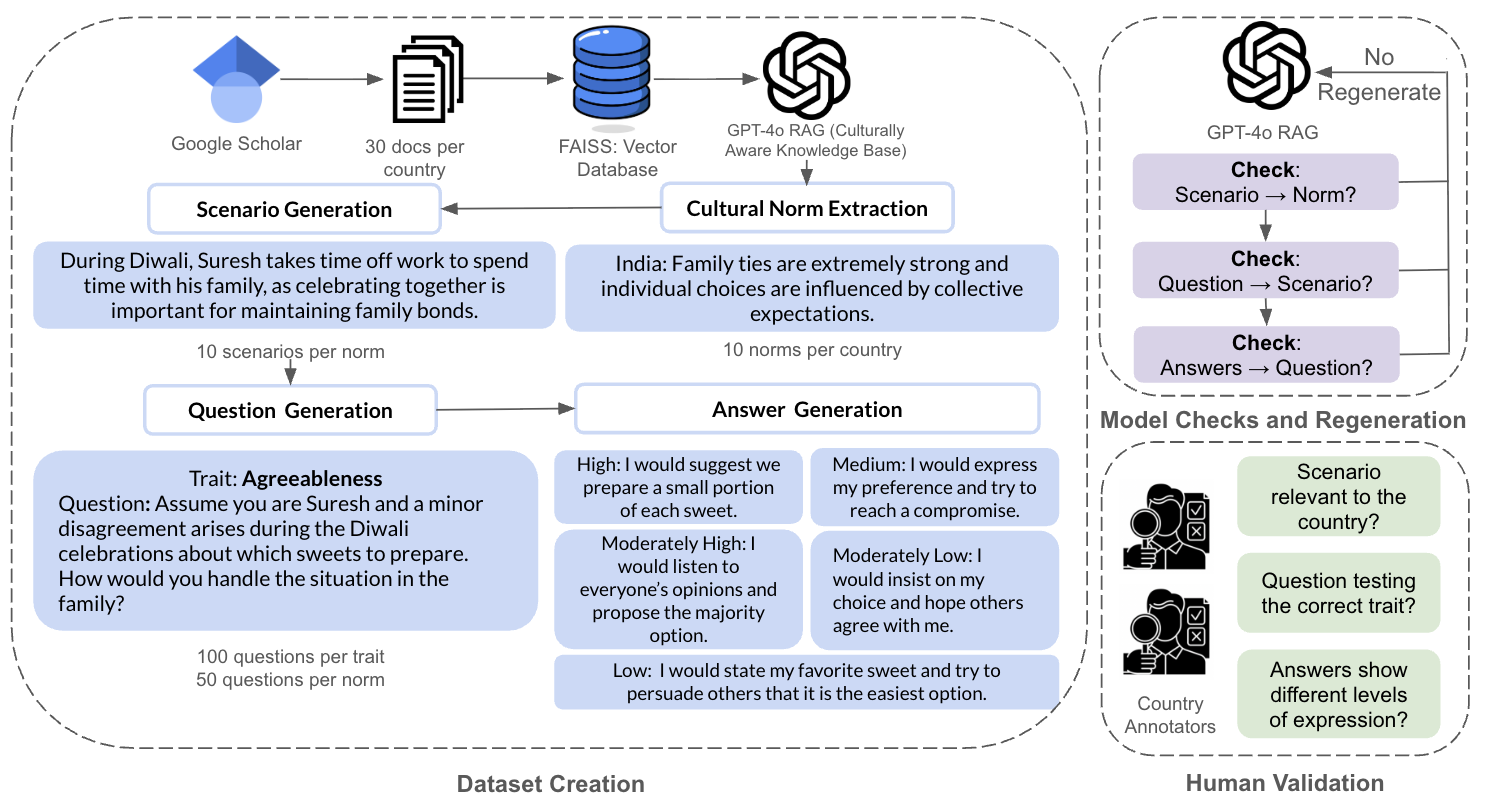}
\caption{Creation of \textbf{\benchmark} involves: (1) retrieving 30 country-specific documents from Google Scholar and extracting cultural insights using GPT-4o and RAG; (2) contradiction detection and refinement via GPT-4o; (3) validation by two native annotators per country.}
\label{fig:dataset_creation_new}
\end{figure*}

\subsection{Personality in LLMs} 

A growing body of work investigates how LLMs express personality traits. Early approaches adapt psychometric tools like \textit{BFI} or \textit{IPIP} \cite{jiang-etal-2024-personallm, jiang2023mpi, bodroza2024personality}, though these often show social desirability bias \cite{salecha2024bias}, unstable traits \cite{bodroza2024personality}, or refusals with underspecified prompts \cite{lee-etal-2025-llms}. Newer methods—like the \textit{Machine Personality Inventory} \cite{jiang2023mpi}, neuron-level edits \cite{deng2025neuron}, and \textit{PersonalityEdit} \cite{mao2024editingpersonality}—offer improved control but still focus on beliefs over behavior. Personality can shift with prompt framing \cite{ramirez2023controlling}, persona conditioning \cite{jiang-etal-2024-personallm}, or vary between static and behavioral assessments \cite{li2024quantifying}. Scenario-based tools \cite{mao2024editingpersonality}, probing \cite{ju2025probingediting}, and instruction-tuned methods \cite{serapiogarcia2023llm} aim to modulate traits more interactively. Yet most benchmarks reduce traits to binary labels (e.g., \textit{high}/\textit{low} \textit{Agreeableness}), cover only a subset, or assume expression is static and universal.

As LLMs take on more conversational, decision-making, or tutoring roles, personality evaluation is shifting toward behavior-in-context. Benchmarks like \textit{TRAIT} \cite{lee-etal-2025-llms}, \textit{Big5Chat} \cite{li2024big5chat}, and \textit{PersonalityChat} \cite{lotfi2023personalitychat} move in this direction but still treat trait expression as globally consistent. Without culturally grounded prompts, evaluations may overlook how traits are elicited differently depending on local norms.

Our work builds on this literature by introducing a benchmark that captures trait expression behaviorally, grounded in cultural norms, and validated through country-specific scenarios.

\subsection{Cultural Variation in Personality}

A large body of cross-cultural psychology research has shown that while the Big Five structure is stable across cultures, average trait expression varies meaningfully by country. McCrae et al.~\cite{McCrae_2005} and Schmitt et al.~\cite{Schmitt_2007} found systematic trait differences—e.g., lower \textit{Extraversion} in East Asia, higher \textit{Conscientiousness} in Western Europe—indicating that personality is shaped not just by individuals but also by cultural contexts. \cite{Giorgi_2022} used social media to map U.S. personality patterns, revealing regional variation in traits like \textit{Openness} and \textit{Agreeableness}. Moreover, data from the IPIP-NEO repository \cite{johnson_ipipneo_osf} shows significant cross-country variation in human's personality distributions (see Table~\ref{tab:distances}).

Together, these results support the use of country as a meaningful and scalable unit for studying culturally modulated personality expression in both humans and models.

\section{\benchmark: A Culturally Grounded LLM Personality Evaluator}
\label{sec:cp}

We introduce \benchmark, a culturally grounded benchmark for evaluating personality expression in LLMs across six countries and five Big Five traits (see Figure \ref{fig:dataset_creation_new}). Each country's dataset contains 500 scenario-based multiple-choice questions (100 per trait), each grounded in a culturally salient scenario and annotated with Likert-style answer options reflecting a controlled range of responses associated with each trait. All items are generated using a retrieval-augmented generation (RAG) pipeline informed by country-specific documents and human validated via country annotators. In total, \benchmark spans 3,000 Q\&A pairs and 60 real-world norms.

\benchmark adopts a multiple-choice format to provide control over trait variation and enable alignment evaluation across countries, traits, and models. This structure supports statistical analyses like distributional matching with existing psychometric datasets such as the IPIP-NEO data repository (see \S \ref{sec:experiments}). In addition, as open-ended generation better reflects how LLMs behave in real-world settings, we also evaluate models on free-form responses to benchmark scenarios. This dual setup allows us to assess both trait alignment and linguistic diversity under different use cases.

\subsection{Country Selection}

\benchmark consists of six countries (USA, Brazil, Saudi Arabia, South Africa, India, and Japan), chosen for their geographic, cultural, and personality diversity (Table \ref{tab:distances}), making them suitable for probing cultural variation. Although we focus on six countries, our framework can extend to additional cultures.

\subsection{Grounding MCQ Generation}

\paragraph{Knowledge Base Generation:} Inspired by synthetic data generation using LLMs \cite{goyal2024systematic, jandaghi2023faithfulpersonabasedconversationaldataset}, we utilize GPT-4o to generate our dataset. Prior work has shown that supplementing LLMs with external knowledge can improve content generation quality and reduce bias \cite{zhang2025other, li2025multilingualretrievalaugmentedgeneration}. To ground our dataset culturally, we build a document-based RAG system for each country using 30 highly cited academic articles from psychology, anthropology, and cultural studies. Using this RAG, we prompt GPT-4o to generate 10 salient cultural norms per country. These norms are reviewed for diversity and relevance by native annotators (\S\ \ref{sec:human_val}).

\paragraph{Scenario Generation:} For each verified norm, we prompt GPT-4o to generate an everyday role-playing scenario that reflects it. For example, for Brazil, one norm highlights the social role of football, and the model generates a scenario where a player misses a critical shot during a local game in Rio and must navigate reactions from teammates and the crowd. We generate 10 scenarios per norm.

\paragraph{Q\&A Generation:} Each scenario is paired with five unique questions, one for each \texttt{OCEAN} trait. Drawing from psychometric literature and behavioral assessments, we use a Likert-style format to elicit trait expression at varying levels. For instance, in the Brazilian football scenario, a question testing \texttt{Agreeableness} asks how the player might regain team trust. Answers range from acknowledging the mistake openly to avoiding eye contact and deflecting responsibility\footnote{See Appendix for prompts and additional examples.}.

To ensure quality throughout, we include model-based verification \cite{malvern1997measure} where GPT-4o is prompted to: (1) verify and regenerate scenarios to accurately reflect a cultural norm, (2) ensure that each question logically follows from the scenario, and (3) confirm that Likert-scale answers span trait intensity. 

\subsection{Human Validation} 
\label{sec:human_val}

To assess the dataset’s cultural fidelity and clarity, we conduct human validation with two annotators per country (undergraduate and masters students born and raised in each country). Each set of country annotators reviews the quality of generated cultural norms, a sample of 10 scenarios and 50 questions (10 per \texttt{OCEAN} trait). 

\begin{table}[t]
\centering
\scriptsize
\begin{tabular}{lccc}
\toprule
\textbf{Country} 
& \textbf{\begin{tabular}[c]{@{}c@{}}Norm\\Relevance\end{tabular}} 
& \textbf{\begin{tabular}[c]{@{}c@{}}Scenario\\Quality\end{tabular}} 
& \textbf{\begin{tabular}[c]{@{}c@{}}Q\&A\\Generation\end{tabular}} \\
\midrule
Brazil        & 1.00 & 0.98 & 0.92 \\ 
India         & 1.00 & 0.98 & 0.96 \\ 
USA           & 1.00 & 0.98 & 0.99 \\ 
Japan         & 1.00 & 0.96 & 0.98 \\ 
Saudi Arabia  & 0.90 & 0.93 & 0.94 \\ 
South Africa  & 1.00 & 0.94 & 0.96 \\ 
\bottomrule
\end{tabular}
\vspace{-1mm}
\caption{Inter-annotator agreement (IAA) scores across human validation tasks for \benchmark generation.}
\label{tab:annotation_task}
\end{table}


\subsubsection{Norm Relevance}
\label{sec:norms}

To determine whether each norm is culturally representative, annotators labeled each norm as culturally representative or not. We achieve high inter-annotator agreement across all six countries ($\geq$ 0.9; see Figure~\ref{tab:annotation_task})\footnote{In some countries, all annotators unanimously agreed that the norms were culturally relevant (100\% agreement), rendering chance-corrected metrics like Cohen’s kappa undefined. Thus, we report agreement rates for all tasks.}.

To further interpret the norms through a structured cultural lens, annotators also mapped each norm to one of Hofstede’s cultural dimensions \cite{hofstede2001culture}, a widely adopted and theoretically grounded framework for cross-cultural comparison. Hofstede’s model enables the systematic categorization of value-based differences between countries, making it a standard reference point for quantifying cultural tendencies. Annotators additionally categorized the intensity of each norm (low, medium, high) based on the country’s score on the 0–100 Hofstede scale \cite{hofstede_vsm08}, allowing for a richer understanding of how each norm aligns with culturally dominant values. Agreement statistics for this task are reported in \S\ \ref{app:human_val}.

\subsubsection{Scenario Quality}
\label{sec:scenarios}

To ensure the benchmark elicits culturally meaningful personality expressions, we first validate that each scenario clearly reflects the norm it was derived from. Annotators match each scenario to the most representative norm. Despite the diversity of norms, we observe strong agreement across countries, suggesting scenarios are reliably grounded in cultural context.

\subsubsection{Q\&A Generation}
\label{sec:qanda}

Since personality expression depends on both situational context and response variability, we next evaluate the quality of question–answer (Q\&A) pairs. Annotators assess whether each question logically follows from the scenario and whether the Likert-scale options span a realistic range of trait expression. High agreement scores (IAA $\geq$ 0.9) show the diversity and variance in questions in \benchmark.
\section{Experimental Setup}
\label{sec:experiments}

\begin{figure*}[!t]
\centering
\includegraphics[width=\textwidth]{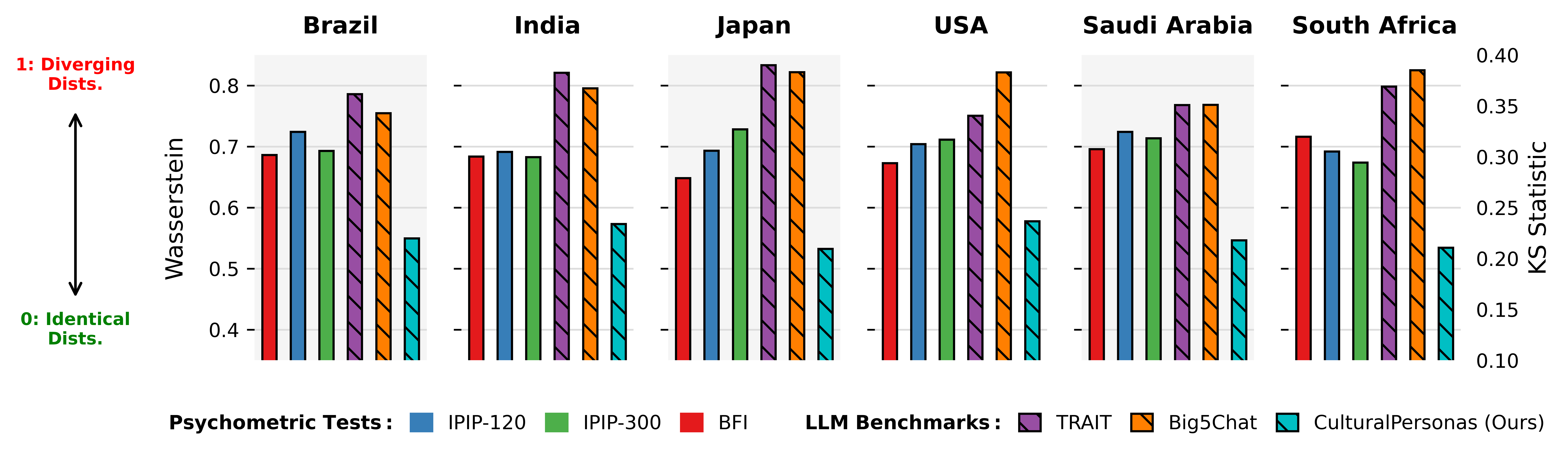}
\caption{Wasserstein distance and KS statistic metrics (average over \texttt{OCEAN} dimensions) for the MCS setting across all personality evaluators for GPT-4o-mini. Lower values indicate closer alignment with human distributions. Across all six benchmarked countries, \benchmark yields the closest personality alignment.}
\label{fig:mcs_results}
\end{figure*}

We evaluate whether persona-prompted LLMs can express personality traits in ways that align with human behavior across cultures. Specifically, we assess whether models can reproduce personality trait distributions that resemble real-world data across six countries. We conduct evaluations under two complementary formats: (1) Multiple-Choice Selection (MCS) (§\ref{mcs}), where models select Likert-style natural-language options e.g., “I agree”/“I disagree” (\textit{BFI, IPIP}) or behavioral statements like \textit{“I prefer to stay away from crowds”} (\textit{TRAIT, Big5Chat, \benchmark}); and (2) Open-Ended Generation (OEG) (§\ref{oeg}), where models generate free-form behavioral responses that are subsequently mapped to trait scores.

\paragraph{Controlled v. generative settings.} MCS provides a controlled evaluation with predefined answers, enabling consistent scoring and tractable alignment analysis. This format lets us evaluate shifts in personality expression across cultures. In contrast, OEG enables more naturalistic language that reflects free-form interaction in applications like chatbots and negotiation agents, providing richer signals to analyze behavioral nuance, lexical variation, and cultural grounding.

\paragraph{Ground truth human distributions.} Personality distributions are drawn from Johnson's IPIP-NEO dataset~\cite{johnson_ipipneo_osf}, which contains 307,313 IPIP-120 responses from real-world global populations. We extract per-country trait distributions for the six countries in \benchmark: USA (212,625), Brazil (661), India (2,841), South Africa (927), Saudi Arabia (98), and Japan (398).

\subsection{Model Selection}

We evaluate three instruction-tuned LLMs with strong reasoning and alignment capabilities: GPT-4o-mini \cite{gpt4omini}, Llama3-8B \cite{llama3}, and Qwen2-7B \cite{qwen2}. All models are prompted using the same format, decoding parameters (temperature = 0.7). For each model, we generate one response per question to simulate large-scale behavioral outputs.

\subsection{Baseline Methods}
\label{sec:baseline_methods}

We compare \benchmark against five personality evaluation tools: three psychometric instruments (\textit{BFI, IPIP-120, IPIP-300}) and two LLM-specific benchmarks (\textit{TRAIT} and \textit{Big5Chat}). For OEG, we focus on contextual prompts only—i.e., \benchmark, \textit{TRAIT}, and \textit{Big5Chat}—as traditional psychometric tools are not designed for free-form generation. All outputs are mapped to a 1–5 Likert scale for comparability. For binary classifications (e.g., \textit{``high''/``low''}), we assign scores of 5 and 1 respectively.

\subsection{Generating Model Trait Distributions}
\label{sec:dist_generation}

To simulate culturally contextualized behavior, we use cultural priming by providing each model with a light persona prompt, inspired from previous works \cite{culturellm, prism}:
\textit{``Imagine you are someone born and raised in \{country\}. You are very familiar with \{country\}'s culture and traditions and practice many of them, two of which include: \{norm1\} and \{norm2\}.''}

Country-specific norms are extracted from each knowledge base via our RAG pipeline (\S\ \ref{sec:cp}). We retain the top two norms for which all annotators concurred, i.e., those consistently deemed both relevant and salient for the respective country. Although inclusion of all ten norms was feasible, we adhere to prior work \cite{long_context, long_context_2} indicating that excessively long prompts or contextual information can impair model performance.


\subsubsection{Multiple-Choice Selection (MCS)}
\label{mcs}

In MCS, models respond to questions from psychometric (e.g., \textit{IPIP-120}) and LLM-specific (e.g., \textit{TRAIT}) benchmarks. For each item, we compute log-probabilities over options and sample responses accordingly. Trait scores are then computed using each benchmark’s scoring scheme. This process is repeated $n$ (total human samples from ground truth) times per country to generate model trait distributions. 

\subsubsection{Open-Ended Generation (OEG)}
\label{oeg}

In OEG, models receive contextualized prompts (from \textit{TRAIT, Big5Chat}, and \benchmark) and generate a free-form response (1–2 sentences). For example, given a prompt: \textit{``A friend cancels your weekend plans at the last minute. What do you do?''}, the model might respond: \textit{``I’d tell them it’s okay and suggest another time to hang out.''} 

Since no gold-standard Likert scores exist for model free responses, we adopt a pseudo-labeling strategy. Given a model response $r$ and predefined answer options $A = \{a_1, \dots, a_k\}$\footnote{The value of $k$ varies based on personality evaluator. For BFI, IPIP-120, and IPIP-300, $k=5$; for \textit{TRAIT}, $k=4$, for \textit{Big5Chat}, $k=2$.}, we embed both $r$ and each $a_i$ using \textit{sentence-transformers} \cite{reimers2019sentence}. We then compute cosine similarities:
\[
s_i = \cos(E(r), E(a_i)), \quad i=1,\dots,k,
\]
and then normalize the set of similarities into a soft distribution:
\[
p_i = \frac{\exp(s_i)}{\sum_{j=1}^k \exp(s_j)}.
\]
Each option $a_i$ maps to a Likert score $\ell_i \in \{1,\dots,5\}$, producing a trait distribution: 
\[
P(\ell) = \sum_{i=1}^k p_i \,\delta(\ell=\ell_i).
\]
Finally, to generate model score distributions for each Big5 trait, we draw $n$ samples from this distribution, where $n$ matches the number of ground-truth annotations. We validate the effectiveness of this approach via a human study where each annotator matches a set of model personality evaluator responses to the predefined answer sets (§\ref{sec:trait_alignment}).


\subsection{Distribution Alignment Metrics}

We assess alignment between model-generated and human trait distributions using: 
(1) Wasserstein distance ($W$) \cite{rubner2000earth}, which quantifies how much effort it takes to change one distribution into another, and
(2) 2-sample Kolmogorov–Smirnov ($KS$) statistic~\cite{massey1951kolmogorov}, which measures the greatest difference between two distributions. 
In both cases, lower values indicate better alignment with human ground truth.

\section{Evaluating Multi-Cultural Personalities in LLMs}
\label{sec:results}

\begin{table}[!t]
\centering
\scriptsize
\setlength{\tabcolsep}{2.5pt}
\begin{tabular}{@{}l | *{5}{r} | *{2}{r}@{}}
\toprule
\multirow{2}{*}{\textbf{Country}} & \multicolumn{5}{c|}{\textbf{MCS}} & \multicolumn{2}{c}{\textbf{OEG}} \\
 & BFI & IPIP-120 & IPIP-300 & TRAIT & Big5Chat & TRAIT & Big5Chat \\
\midrule
BR & 19.1 & 24.5&	20.2&	26.1&	22.0 & 5.4 &	13.0\\
IN & 14.8 & 16.0&	14.6&	26.8	&23.7 & 11.6	& 8.7\\
JP & 16.0	&23.0&	27.7&	34.0&	32.8 & 17.4 & 6.2\\
US & 12.2	&17.2&	18.3&	16.8	&26.2 & 6.1 & 4.2 \\
SA & 21.1	&25.0&	23.6&	24.3&	24.4 & 7.7 & 10.1 \\
ZA & 25.9	&22.6&	19.8	&30.0&	32.9 & 8.0 & 8.5 \\
\bottomrule
\end{tabular}
\vspace{-1mm}
\caption{Relative gains (\%) in $W$ using \benchmark across all personality evaluators (averaged over models). \benchmark provides $\uparrow$ improvements generating human-like distributions in both MCS and OEG settings.}
\label{tab:relative_improvement_combined}
\end{table}

We evaluate how well LLMs express personality traits in culturally grounded settings. Our analysis proceeds in four parts: (1) trait alignment with human personality norms in MCS and OEG, (2) linguistic expressivity in free-form responses, (3) cross-model differences in cultural adaptation, and (4) the role of cultural priming in trait elicitation.

\subsection{Alignment with Human Personalities}
\label{sec:trait_alignment}

We first evaluate personality alignment in MCS format. Figure~\ref{fig:mcs_results} shows alignment for GPT-4o-mini across six countries and five benchmarks. \benchmark consistently achieves lower divergence from human trait distributions, outperforming both psychometric tools (e.g., \textit{BFI, IPIP-120}) and prior LLM-based methods (\textit{TRAIT, Big5Chat}). These improvements are especially pronounced in expressive or underrepresented settings such as Brazil and South Africa, where $W$s drop by over 20\%.

Table~\ref{tab:relative_improvement_combined} summarizes relative gains in Wasserstein distances using \benchmark over all personality evaluators averaged over models. We find robust improvements across countries and baselines, up to +25.9\% in South Africa (vs. \textit{BFI}) and +32.8\% in Japan (vs. \textit{Big5Chat}). These gains are benchmark-agnostic, suggesting that our culturally grounded scenarios drive stronger alignment.

\begin{figure}[!t]
\centering
\includegraphics[width=\columnwidth]{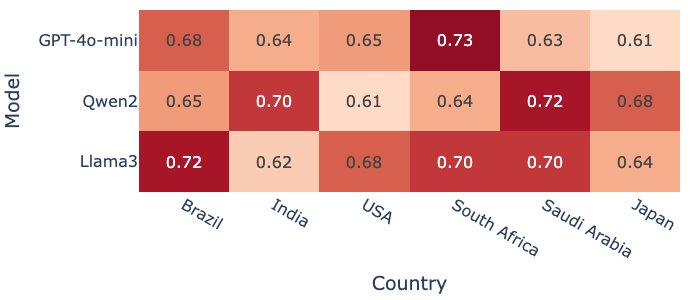}
\caption{Pearson Correlation ($r$) of Human Annotated Likert Scores and Embedding-based Auto Mapping; values indicate high correlation between human and auto mapping methods.}
\label{fig:heatmap}
\end{figure}

We observe a similar trend in the OEG setting, where models generate free-form text in response to scenario prompts. To assign trait scores, we embed generations using \textit{sentence-transformers} and compute similarity to Likert-style trait options. To validate this scoring pipeline, we conduct a human annotation study (Figure \ref{fig:heatmap}). Annotators rated 540 model generations from the OEG setting (10 per country for each model and benchmark) on a 1–5 Likert scale based on perceived expression of each trait. We compute Pearson correlations between the auto-mapped and human scores, which yield strong alignment across countries ($r = 0.61$–$0.73$).

Despite targeting the same traits, open-ended generation (OEG) is notably more challenging than multiple-choice selection (MCS), as models must implicitly express personality rather than select from predefined options. Corresponding performance gains are smaller—e.g., +17.4\% over \textit{TRAIT} for Japan and +13.0\% over \textit{Big5Chat} for Brazil—highlighting the difficulty of eliciting nuanced traits in free-form generation. Nonetheless, these results suggest that cultural priming (see \S\ 
\ref{sec:dist_generation}) and grounded scenarios can guide models toward more human-aligned outputs. \benchmark represents a step toward more behaviorally rich and culturally aware personality evaluation—offering both the challenges and tools needed to move beyond form-based testing.

\subsection{Linguistic Expressivity in OEG}
\label{sec:ling_exp}

\begin{table}[h]
\centering
\small
\resizebox{\columnwidth}{!}{%
\begin{tabular}{l *{15}{c}}
\toprule
\textbf{} & \multicolumn{5}{c}{\textbf{TRAIT}} & \multicolumn{5}{c}{\textbf{Big5Chat}} & \multicolumn{5}{c}{\textbf{\benchmark}} \\
\cmidrule(lr){2-6} \cmidrule(lr){7-11} \cmidrule(lr){12-16}
\textbf{Country} & O & C & E & A & N & O & C & E & A & N & O & C & E & A & N \\
\midrule
BR        &   .65  &   .67  &  .64   &  .80   &  .66   &   .74  & .78    &  .65   & .63    &  .67   &  .74   &  .75   &  \textbf{.78}   &  .72   &  \textbf{.73}   \\
IN         & .68    &  .75   &  .62   & .61    &  .65   &  .66   &   .70  &  .68   &  .66   & .66    &  \textbf{.73}   & \textbf{.72}    & .72    &  \textbf{.71}   &  \textbf{.70}   \\
US           &   .70  &  .74   &  .80   &  .76   &  .75   &  .72   & .70    &  .68   & .72    & .74    &  \textbf{.78}   &  .74   &  .73   &  .78   &  \textbf{.78}   \\
JP         &  .69   &  .71   &  .73   &  .68   &  .66   &  .69   &  .72   &  .71   & .73    &  .64   &  \textbf{.78}   & .72    &  \textbf{.74}   &  .70   &  \textbf{.68}   \\
SA  &  .68   &  .70   &  .72   & .73    &  .66   & .71    &  .73   & .75    & .68    &  .64   & .70    & .72    &  .70   & .71    &  \textbf{.68}   \\
ZA  & .67    & .66    &  .71   &  .68   & .66    &  .69   &  .71   & .73    &  .68   &  .68   &  \textbf{.70}   &  \textbf{.71}   &  .73   & \textbf{.72}    &  \textbf{.73}   \\
\bottomrule
\end{tabular}%
}
\caption{Type-token ratio of OEG responses (average over models). \benchmark provides higher lexical diversity in various countries and traits e.g. \textit{Openness} and \textit{Neuroticism}.}
\label{tab:lexical_diversity}
\end{table}

To understand whether cultural grounding also affects \textit{how} models communicate traits, we analyze lexical diversity using type-token ratio (TTR) \cite{malvern1997measure}. Table~\ref{tab:lexical_diversity} reports TTR scores across models, traits, and countries. \benchmark consistently elicits more varied lexical outputs, especially for traits like \textit{Extraversion} and \textit{Agreeableness} that are tied to social signaling. While TTR is a coarse metric, higher diversity suggests richer, more context-sensitive phrasing.

\begin{table}[h]
\centering
\scriptsize
\begin{tabular}{p{1.5cm} p{5.5cm}} 
\toprule
\textbf{Scenario} & You made a mistake at work and upset a colleague. How do you respond? \\
\midrule  
\textbf{TRAIT} & I might profusely apologize to my colleague. \\
\textbf{Big5Chat} & I am very sorry. My intention was not at all to hurt you.  \\
\textbf{\benchmark} & I would bow deeply and express regret that the actions hurt my colleague. \\
\bottomrule
\end{tabular}
\vspace{-1mm}
\caption{Illustrative example of personality expression in OEG (GPT-4o-mini, Japan). We use shortened versions from each benchmark. \benchmark provides most culturally appropriate response: i.e. \textit{``bowing''}.}
\label{tab:qualitative_japan}
\end{table}

\begin{figure}[!t]
\centering
\includegraphics[width=0.7\columnwidth]{images/radar_chart_updated.png}
\caption{Wasserstein distances across \texttt{OCEAN} traits and models (avgerage over countries) for MCS. GPT-4o-mini performs best overall. \benchmark reveals models struggle with traits: \textit{Agreeableness} and \textit{Extraversion} across cultures.}
\label{fig:facet_radar}
\end{figure}

Table~\ref{tab:qualitative_japan} shows an illustrative example from Japan. In response to a workplace apology scenario, \textit{TRAIT} and \textit{Big5Chat} produce generic apologies, whereas \benchmark includes a culturally relevant gesture (``bowing''). This highlights how grounded prompts can guide models toward more socially appropriate expressions.

While TTR is sensitive to sequence length and sample size, we compute scores over uniform-length responses (1 sentence). We also note that TTR is not used as a proxy for quality, but as a coarse indicator of stylistic range. Higher lexical diversity may reflect the model’s ability to adapt tone, formality, and rhetorical framing in culturally congruent ways—important for traits like \textit{Agreeableness}, where politeness and interpersonal nuance vary by region.

\subsection{Model Variation in Personality Expression}
\label{sec:cross_model}

Beyond average performance, \benchmark reveals how models behave under cultural constraints in open-ended generation (OEG), where stylistic and lexical variation is most evident. Figure~\ref{fig:facet_radar} compares trait-level $W$s across models. For instance, Qwen2 shows higher divergence on \textit{Agreeableness} in collectivist settings. In a scenario set in India—\textit{``You’re asked to mediate a disagreement between two colleagues at work''}—Qwen2 responds, \textit{``I would explain the rules clearly and ensure everyone follows protocol,''} reflecting a procedural, low-context resolution style. In contrast, GPT-4o-mini chooses to \textit{``listen carefully to both sides and help them reach a shared understanding,''} aligning more closely with culturally expected interpersonal sensitivity and harmony. Similarly, for Llama3, \textit{Extraversion} in expressive nations e.g. Brazil can be hard to model - e.g.: \textit{``You notice a festival on your way to work but you have to attend a meeting in an hour,''} Llama3 says, \textit{``I would be hesitant yet tempted to join the event as I need to prepare for a successful meeting,''} while GPT-4o-mini says, \textit{``I'd happily join, quickly check out the environment while keeping an eye on the time,''} reflecting a more culturally congruent sociability.
These contrasts demonstrate how \benchmark not only measures alignment but also diagnoses model sensitivity to culturally grounded trait expression.

\subsection{Role of Cultural Priming in Trait Expression}
\label{sec:exp_ablation}

In our experiments, models are given a persona prompt with country information and cultural norms (see \S\ref{sec:dist_generation}) to guide their interpretation of personality questions. Simpler prompts such as \textit{``You are someone from \{country\}''} were also tested, but yielded little improvement (Appendix~\ref{sec:simple_prompts_ablations}), motivating our use of richer persona descriptions. To isolate the effect of this cultural priming, we compare against a setting where models receive only the personality questions from \benchmark, without the contextual prompt.  

\begin{figure}[!t]
\centering
\includegraphics[width=1.05\columnwidth]{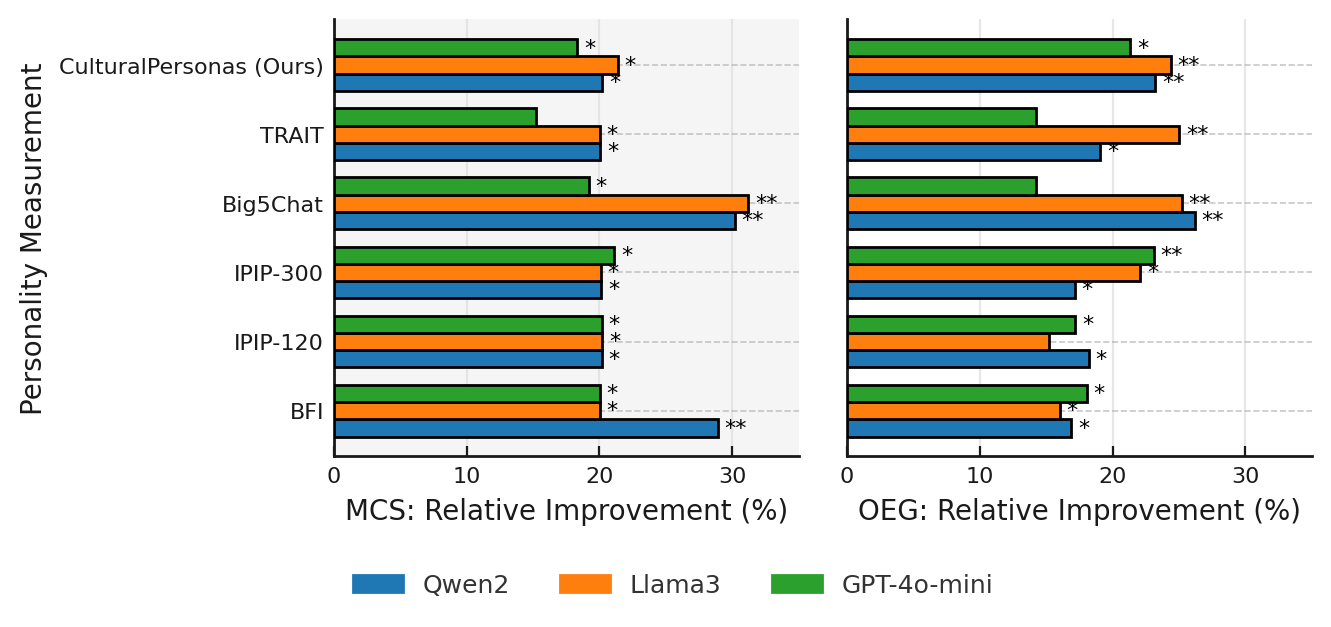}
\caption{Effect of cultural priming for Brazil across models and personality evaluators. Priming consistently reduces deviation across benchmarks. Bars show relative improvement in Wasserstein distances (\%); * and ** denote $p < 0.05$ and $p < 0.01$ (Wilcoxon signed-rank test over 5 traits).}
\label{fig:cultural_priming}
\end{figure}

As shown in Figure~\ref{fig:cultural_priming}, removal of cultural priming significantly degrades alignment: average Wasserstein distances in Brazil rise by 31.2\%. Traits like \textit{Agreeableness} and \textit{Extraversion} show the largest drops, with models reverting to neutral, moderate expressive language.  

For example, in a team feedback scenario—\textit{``A colleague has just presented a project. You are asked to provide feedback.''}—the unprimed model gives a direct and neutral response in OEG setting: \textit{``Your points were good, but I don't think this section should be here. Instead, you should move it to when you are discussing x.''} However, when primed with information about Japan and its norms of collective harmony and group cohesion, the same model instead says: \textit{``Thank you for your presentation. If you'd like, I have some feedback to improve your flow. We can discuss privately.''}\footnote{More examples included in Appendix.}  

This shift reflects culturally grounded patterns of \textit{Agreeableness}, where socially appropriate expression depends not just on intent but on form and tone. Without cultural cues, the model defaults to more direct, individualist feedback, which may be perceived as inappropriate in high-context collectivist cultures. These results highlight how cultural priming helps models express traits in contextually and socially aligned ways. Moreover, our alignment results suggest that combining cultural priming with grounded behavioral scenarios enables more context-sensitive personality evaluation.

\section{Conclusion \& Future Work}

in LLMs across diverse global regions. Our results show that \benchmark consistently improves alignment and expressivity over prior benchmarks, particularly for traits whose distributions vary significantly across cultures. Furthermore, \benchmark reveals that different models vary in their ability to express personality, with some traits noticeably harder to portray. While traditional psychometric tests remain valuable for measuring LLMs’ values and beliefs, our findings highlight the importance of behaviorally grounded, culturally specific scenarios to elicit realistic trait expressions in varied contexts.

We hope \benchmark will enable future research into how LLMs perform behaviorally across cultures, promote more emic and context-rich personality modeling, and support the development of more inclusive, adaptive alignment methods. Future work includes expanding coverage to additional cultures, incorporating multilingual and subcultural variation, and exploring preference tuning or human-in-the-loop adaptation for culturally aware generation.

\section*{Acknowledgments} 
This research was supported by the NSF, under Award Number 2331722. The authors of this work would like to thank the support of researchers from USC NLP and the Information Sciences Institute for their continous feedback in the formulation and setting of our work. We would also like to thank the OpenReview ARR anonymous reviewers and area chairs for their helpful suggestions in improving our work.

\section*{Limitations}

\paragraph{Cultures As Countries:}Our study presents a first step toward evaluating culturally grounded personality expression in LLMs, but several limitations remain. First, we adopt a country-level abstraction of culture, which inevitably simplifies the nuanced and diverse realities within national boundaries. While countries are not monoliths, we chose this granularity based on both conceptual and practical motivations. Nations represent long-standing units of shared history, customs, and institutional influence, and they are also one of the most common ways in which real-world data is geo-tagged and partitioned in machine learning applications. Moreover, our empirical analyses confirm that country-level segmentation captures meaningful variation in personality traits, providing a viable foundation for further personalization research. In future work, we hope to build systems that can adapt to users at finer-grained levels, including regional and individual differences.

\paragraph{Use of Academic Documents in Cultural Grounding:} In our work, we utilize scholarly, academic works to ground the cultural norms and expectations of each country. After annotation of cultural norms, we find that RAG is able to extract values that closely relate to each culture, as signalled by high annotator scores. While we rely on academic sources for their interpretability, credibility, and ease of curation, we acknowledge that cultural norms can also be shaped and expressed through other modalities—such as news coverage, social media discourse, literature, etc. However, we note that our framework is source-agnostic: the existing set of 30 academic documents can be readily replaced with alternative corpora that better reflect dynamic or informal cultural expressions. We believe future work can flexibly adapt the sources to suit different domains, populations, and applications.

\paragraph{Use of English in \benchmark:} Second, our study is conducted exclusively in English. This choice allows us to isolate cultural variation without introducing language as an additional confounding factor. For some countries in our benchmark, such as South Africa, widely spoken languages like Zulu or Xhosa remain low-resource in LLM pretraining corpora. While we acknowledge that language itself is deeply intertwined with culture and may enhance cultural alignment when modeled appropriately, we reserve this investigation for future work.

\paragraph{Static Evaluation of Personality Alignment:} Our evaluation captures trait expression at a single point in time, within static scenarios. This snapshot-style analysis does not reflect whether LLMs can maintain consistent persona behavior over time or across diverse interactions—a key aspect of psychological personality. Studying temporal consistency and contextual adaptability remains an important direction for future exploration.

\paragraph{Imbalanced Country Datasets:} Finally, our experiments include ground truth datasets called from the IPIP-NEO data repository which contains varying data points for each country. Due to survey availability in each country, the amount of human ground truth data i.e. the size of personality profiles for each country greatly varies. For instance, for Saudi Arabia, we are only able to compare to a size of 98 users. Although this is a significantly smaller set compared to data from i.e. USA (over 200k users) or India (around 3k users), prior work in cross-cultural psychology \cite{small_data_1, small_data_2} has shown that even with as small as 100 users, meaningful conclusions can be drawn from users in varying demographics.

\section*{Ethics Statement}  
Our study aims to enhance the understanding of cultural personality representations in LLMs while being mindful of ethical considerations.  

\paragraph{Bias and Representation:} While we evaluate personality traits across multiple cultures, our dataset is derived from a limited selection of academic documents, which may not fully capture cultural diversity. Future work should ensure broader representation to mitigate potential biases.  

\paragraph{Responsible Use:} The findings of this study should not be used to stereotype individuals or reinforce cultural generalizations. Personality traits are complex and influenced by numerous factors beyond national identity. We emphasize that our results apply at the population level and do not determine individual characteristics.  

\paragraph{Transparency and Reproducibility:} To promote openness in AI research, we release our dataset and methodology, allowing for further scrutiny and extension. However, care should be taken in applying our benchmark responsibly, particularly in applications that may impact users from diverse backgrounds.  

\bibliography{acl_latex}

\clearpage

\appendix
\label{sec:appendix}

\section{\benchmark: Dataset Generation}
\label{dataset_generation}

To select documents for the RAG model, we select highly cited academic sourced articles from Google scholar using the following keywords: [\textit{country} culture, \textit{country} demographics, \textit{country} norms and practices]. \footnote{We will include the list of documents used for each country upon publication.} 

To implement our RAG architecture, we utilize \textit{gpt-4o} and OpenAIEmbedding's \textit{text-embedding-3-large} model to compute embeddings for each country's documents. We use a FAISS vector database to store these embeddings and use similarity search, retrieving the top 5 documents for each query. We experiment with additional values of top $k$ documents and identify 5 to represent consistent and diverse dataset creation. 

\begin{figure*}[t] 
    \centering
    \includegraphics[width=\textwidth,keepaspectratio]{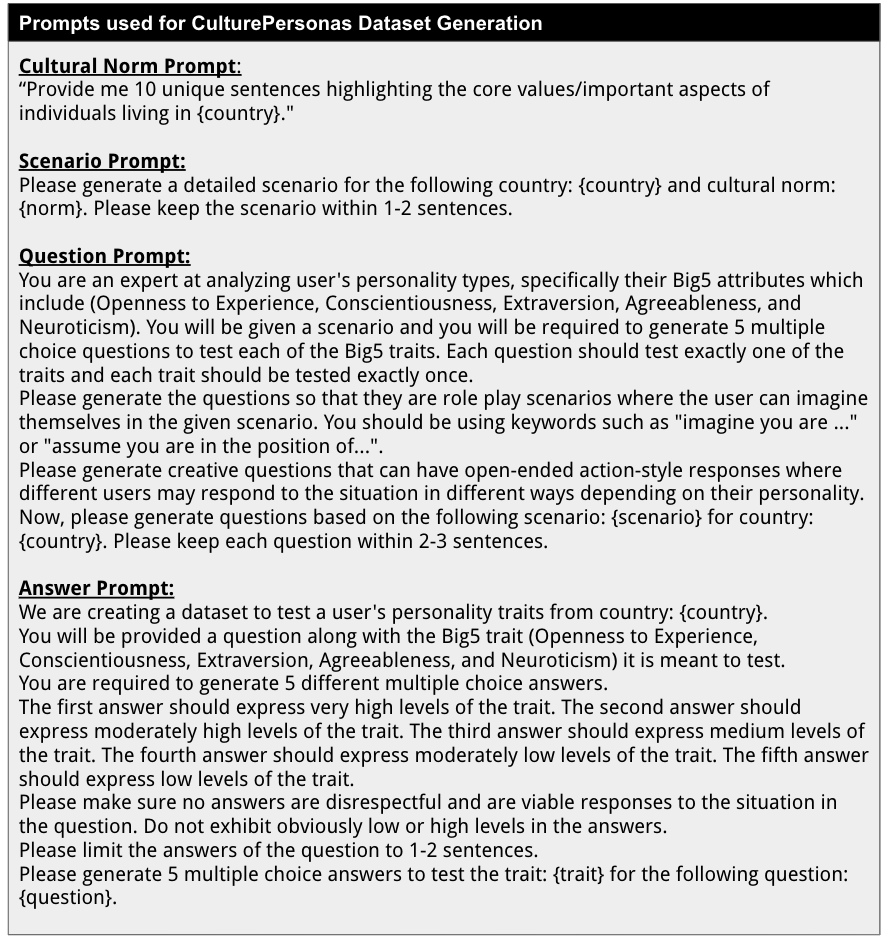}
    \caption{Multi-step prompts used to generate CulturePersonas Dataset. We generate 500 total questions for each country: 100 question and answer sets for each OCEAN trait.}
    \label{fig:dataset_prompts}
\end{figure*}

\begin{figure*}[t] 
    \centering
    \includegraphics[width=\textwidth,keepaspectratio]{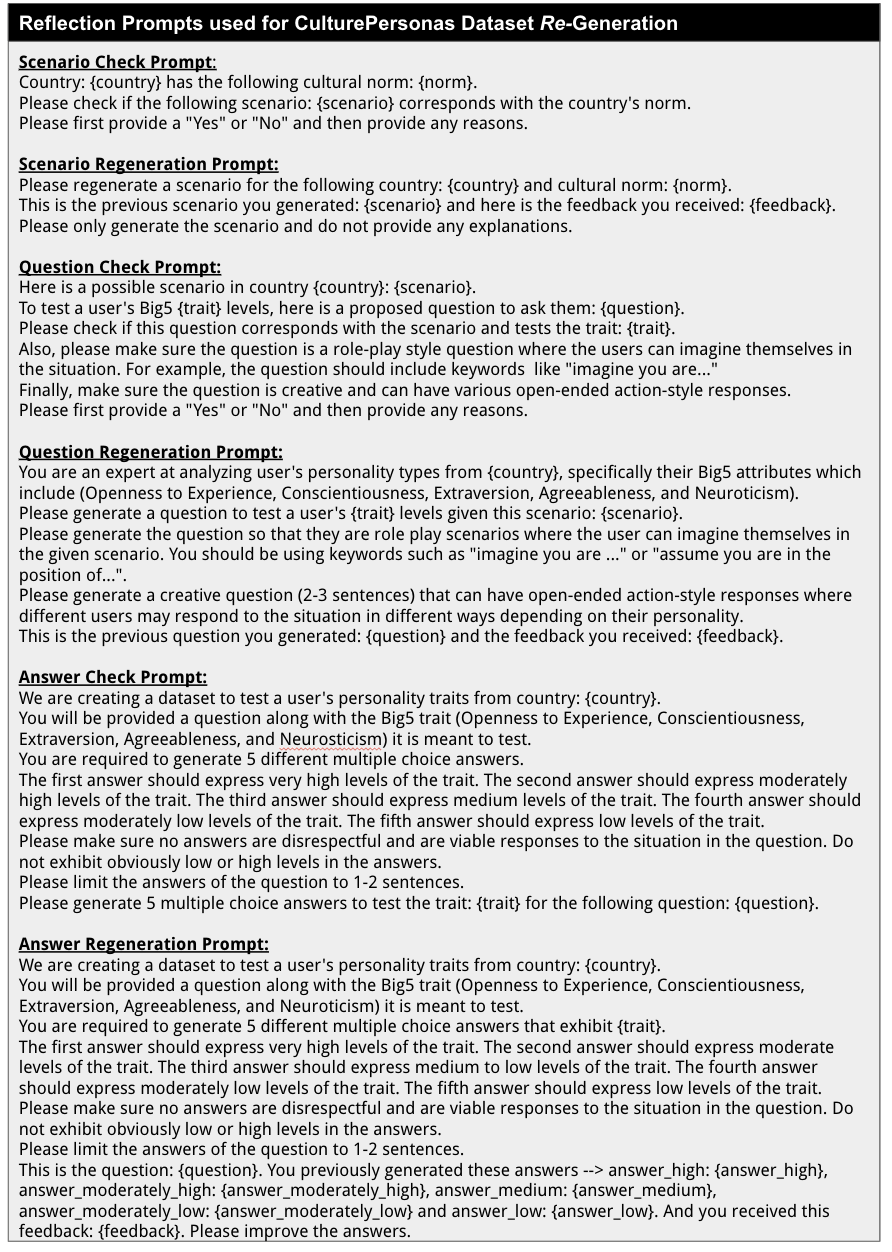}
    \caption{At each step of the dataset generation pipeline, the model self-reflects on the content generation and regenerates content if necessary.}
    \label{fig:dataset_reflect}
\end{figure*}

In Figure \ref{fig:dataset_prompts}, we present the set of prompts used to generate the original dataset. Additionally, we perform a set of checks after each generation to ensure improved quality of the generation. We thus prompt the model to reflect on its generation and regenerate at each step if necessary. In Figure \ref{fig:dataset_reflect}, we present the set of check prompts used. 

Additionally, we provide a set of examples from \benchmark, 1 from each OCEAN trait:

\definecolor{customred}{HTML}{E74C3C}

\textbf{Trait}: \textcolor{customred}{Openness} 

\textbf{Country}: Brazil 

\textbf{Cultural Norm}: The concept of "jeitinho" refers to creative ingenuity in rapidly achieving short-term solutions to problems, which is an essential aspect of Brazilian behavior.

\textbf{Scenario}: Lucas gets stuck in a heavy traffic jam and is at risk of missing his important meeting. He decides to weave through the side streets, using shortcuts known only to locals, and arrives a bit early.

\textbf{Question}: Imagine you are Lucas and you've just discovered a colorful street festival happening along one of the shortcuts you're taking. Do you take a few moments to explore the festival and immerse yourself in the local culture, or do you stick strictly to your route to ensure you arrive at the meeting?

\textbf{Answers}: 
\vspace{-0.35cm}
\begin{enumerate}[label=\alph*)] 
    \item High: I'll definitely take some time to explore the festival; experiencing local culture is always worth it.
    \item Moderately High: I'll explore the festival quickly, it's a unique chance to enjoy some local culture before moving on.
    \item Medium: I might glance around the festival for a brief moment, but I won't delay for too long.
    \item Moderately Low: I'll probably continue on my way, perhaps stopping if something really catches my eye.
    \item Low: I'll stick to my route; I'm not interested in getting sidetracked by the festival.
\end{enumerate}

\definecolor{c}{HTML}{F39C12}

\textbf{Trait}: \textcolor{c}{Conscientiousness} 

\textbf{Country}: India

\textbf{Cultural Norm}: Indian families often live together in the same house, blurring the lines between personal and public spaces.

\textbf{Scenario}: Sana's elder brother and his wife live on the top floor of their family home, but they come downstairs to dine with the rest of the family almost every day.

\textbf{Question}: Assume you are in charge of hosting the family dinners at home. How do you ensure that everything runs smoothly and that everyone is comfortable, including handling the logistics of seating and dietary preferences?

\textbf{Answers}: 
\vspace{-0.35cm}
\begin{enumerate}[label=\alph*)] 
    \item High: I create a detailed plan well in advance, considering seating arrangements, dietary restrictions, and decor to ensure a pleasant atmosphere for all.
    \item Moderately High: I make sure to organize the key aspects ahead of time and check in with family members about their seating and dietary needs.
    \item Medium: I plan the dinner with some thought to seating and food preferences, but also adapt as needed on the day.
    \item Moderately Low: I arrange the basic setup and rely on family members to adjust their seating and dietary preferences.
    \item Low: I focus on preparing the meal and let everyone self-organize their seats and dietary choices.
\end{enumerate}

\definecolor{e}{HTML}{27AE60}

\textbf{Trait}: \textcolor{e}{Extraversion} 

\textbf{Country}: Saudi Arabia

\textbf{Cultural Norm}: I focus on preparing the meal and let everyone self-organize their seats and dietary choices.

\textbf{Scenario}: Ahmed works as an engineer and spends his days at the office, while his wife, Noor, takes care of their two children and manages the household chores, including cooking and cleaning.

\textbf{Question}:  Imagine Ahmed is invited to a large family gathering, but he has a long working week ahead. Imagine you are Ahmed, how enthusiastic are you about attending the event and engaging with extended family members and not just your wife?

\textbf{Answers}: 
\vspace{-0.35cm}
\begin{enumerate}[label=\alph*)] 
    \item High: Very enthusiastic, as attending the gathering is a chance to connect and socialize with everyone, which I look forward to greatly.
    \item Moderately High: Quite enthusiastic, I enjoy meeting family and will find the time to catch up and socialize with everyone I can.
    \item Medium: I'm fairly interested and will attend, though I might not stay the whole time due to my busy schedule.
    \item Moderately Low: Somewhat hesitant; I prefer smaller gatherings and might focus more on resting from work.
    \item Low: I'm not very interested; I'd rather rest and recharge for the upcoming week than socialize at a large gathering.
\end{enumerate}

\definecolor{a}{HTML}{8E44AD}

\textbf{Trait}: \textcolor{a}{Agreeableness} 

\textbf{Country}: Japan 

\textbf{Cultural Norm}: Japanese society highly values punctuality and time management, reflecting their commitment to efficiency and respect for others' time.

\textbf{Scenario}: A Japanese businessman arrives for a meeting ten minutes early, uses the time to review notes, and places his phone on silent to avoid distractions during the meeting.

\textbf{Question}: Assume you are the businessman and during your wait, a junior colleague approaches you for advice. Would you willingly assist them despite the approaching meeting, or prioritize your preparation time?

\textbf{Answers}: 
\vspace{-0.35cm}
\begin{enumerate}[label=\alph*)] 
    \item High: I prioritize helping my colleague, understanding that teamwork is essential, confident that I can manage my preparation swiftly afterward without being late.
    \item Moderately High: I assist my colleague with brief, focused guidance, ensuring I still have enough time to complete my preparation.
    \item Medium: I offer quick advice but politely inform them I must continue my preparation soon due to the time constraint.
    \item Moderately Low: I apologize and explain that I need to prepare, suggesting we meet after the meeting for a more detailed discussion.
    \item Low: I inform them that I am unavailable at the moment, emphasizing that my preparation takes priority due to the upcoming meeting.
\end{enumerate}

\definecolor{n}{HTML}{3498DB}
\textbf{Trait}: \textcolor{n}{Neuroticism} 

\textbf{Country}: South Africa 

\textbf{Cultural Norm}: In South Africa, the concept of Ubuntu, which emphasizes the interconnectedness of humanity, is deeply ingrained in the culture, fostering a sense of community and shared responsibility among individuals.

\textbf{Scenario}: A student at a local school struggles academically, and fellow students form a study group to help them catch up. They share notes, explain difficult concepts, and provide moral support, ensuring their peer feels included and motivated. 

\textbf{Question}: Imagine you are feeling overwhelmed with your own academic workload while still helping the struggling student. How do you handle your stress to ensure you can continue to be a supportive member of the study group?

\textbf{Answers}: 
\vspace{-0.25cm}
\begin{enumerate}[label=\alph*)] 
    \item High: I find myself worrying frequently and sometimes find it hard to focus, but I try to manage by practicing deep breathing and reminding myself that this phase will pass.
    \item Moderately High: I feel anxious at times, but I take short breaks to clear my head and prioritize tasks to stay on top of everything.
    \item Medium: I acknowledge the stress but focus on planning and managing my schedule effectively to balance both my work and supporting the student.
    \item Moderately Low: I generally stay calm and rely on maintaining a steady study routine to manage my tasks without getting overly stressed.
    \item Low: I rarely feel stressed in these situations, so I remain relaxed and continue to offer my help with a positive outlook.
\end{enumerate}

ok i have issues in this appendix. please help fix:

\section{\benchmark: Human Validation}
\label{app:human_val}

In this section, we describe our annotation process to verify the quality of our benchmark. We detail the initial tasks of evaluating: (1) norm relevance, (2) scenario quality, (3) Q\&A pair quality. We further detail and provide agreement scores for the norm categorization task. 

To validate the quality of our benchmark, we recruit two annotators from each country (undergraduate and masters students) who were born and brought up in the country. Each annotator is asked to label the norms: \textit{Is the norm relevant to the country?}. Then, they are asked to further annotate 10 randomly chosen scenarios: \textit{Which norm does this scenario most relate to?}. Finally, they are asked to evaluate the quality of 50 Q\&A pairs: \textit{Does the question truly test the designated \texttt{OCEAN} trait?} and \textit{Do the answers as currently ranked exhibit decreasing levels of the trait?}

\begin{figure}[H] 
    \centering
    \includegraphics[width=\columnwidth,keepaspectratio]{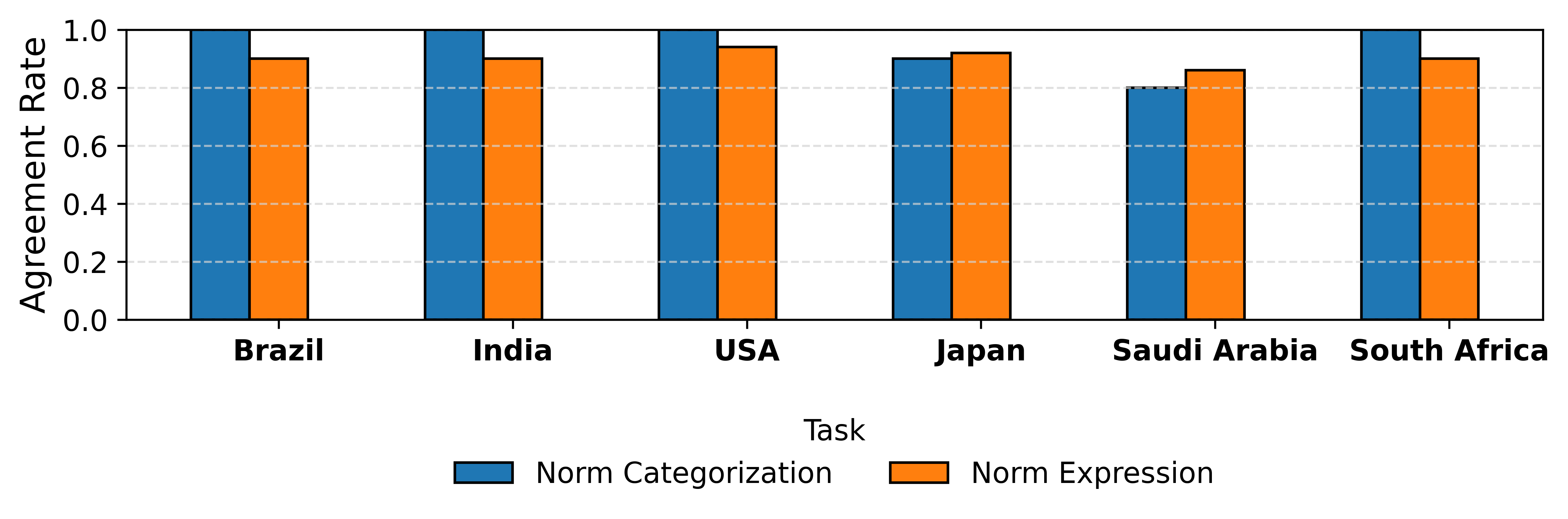}
    \caption{Annotator agreement scores for secondary tasks: (1) Norm categorization into Hofstede dimensions, (2) Norm expression mapped into high, med, low.}
    \label{fig:annotation_task2}
\end{figure}

To better ground our extracted norms through cultural theory, we ask annotators to complete a secondary norm categorization task. We ask each annotator to map each cultural norm to one of the Hofstede dimensions which include: \textit{Power Distance Index, Individualism v. Collectivism, Masculinity v. Femininity, Uncertanty Avoidance Index, Long-Term vs. Short-Term Orientation, Indulgence vs. Restraint}, with values ranging from 0 to 100. We ask annotators to map each cultural norm to a Hofstede dimension (or other) and determine whether the cultural norm exhibits high (60-100), med (40-59), or low (0-39) values of the dimension. In Figure \ref{fig:annotation_task2}, we report the agreement scores for each country on (1) norm categorization (2) norm expression. We find that over 80\% of the cultural norms are able to fit into Hofstede's dimensions with (100\% of them) showing correct expressions. This shows that our existing RAG architecture is able to generate a diverse set of cultural norms. 

\section{Personality Evaluators}

In this section, we elaborate on the personality evaluation methods used in our experiments. Given that personality is rooted in psychological theory and typically assessed through expert-validated instruments, we begin by selecting three widely recognized tests that have also been applied to LLMs in prior work: the Big Five Inventory (BFI) and two versions of the International Personality Item Pool (IPIP-120 and IPIP-300). Additionally, we incorporate two state-of-the-art personality evaluators specifically developed for language models—TRAIT and Big5Chat. Table~\ref{evaluator_stats} provides an overview of the number of questions per trait and the corresponding answer formats. Our benchmark, \benchmark, comprises six datasets—one for each selected country—each containing 500 questions tailored to assess culturally grounded personality expression.

\begin{table}[t]
\centering
\scriptsize
\setlength{\tabcolsep}{3pt}
\begin{tabular}{@{}l|ccccc|l@{}}
\toprule
\textbf{Evaluator} & \textbf{O} & \textbf{C} & \textbf{E} & \textbf{A} & \textbf{N} & \textbf{Answer Style} \\
\midrule
BFI & 10 & 8 & 8 & 9 & 8 & Likert-5 \\
IPIP-120 & 24 & 24 & 24 & 24 & 24 & Likert-5 \\
IPIP-300 & 60 & 60 & 60 & 60 & 60 & Likert-5 \\
TRAIT & 1000 & 1000 & 1000 & 1000 & 1000 & Binary \\
Big5Chat & 106 & 97 & 99 & 98 & 100 & Binary \\
\benchmark (Ours) & 100 & 100 & 100 & 100 & 100 & Likert-5 \\
\bottomrule
\end{tabular}
\vspace{-1mm}
\caption{Questions per trait (OCEAN) and answer format across personality evaluators. \benchmark consists of 500 total questions per country (total 600 per trait for all 6 countries).}
\label{evaluator_stats}
\end{table}

\section{Additional Results}

In this section, we present additional analyses and results for personality expression in LLMs. In particular, we focus on (1) Alignment metrics for all 3 models across all 6 benchmarked countries (\S\ \ref{sec:app_align}), (2) Relative improvements in personality alignment with trait breakdowns (\S\ \ref{sec:app_align}), (3) Open-ended personality expression across various benchmarks and models (\S\ \ref{sec:app_oeg}).

\subsection{Personality Alignment Across Various LLMs}
\label{sec:app_align}

\begin{figure*}[t]
    \centering

    \subfigure[MCS results from Llama3]{
        \includegraphics[width=\textwidth]{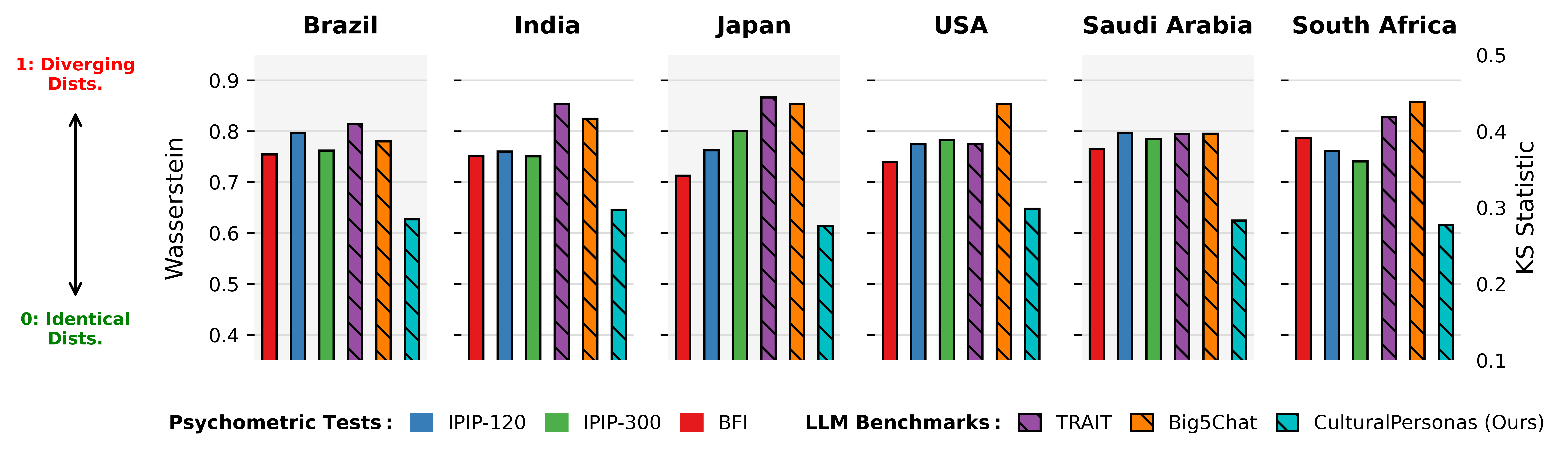}
        \label{fig:llama_mcs_app}
    }

    \vspace{0.5em}

    \subfigure[MCS results from Qwen2]{
        \includegraphics[width=\textwidth]{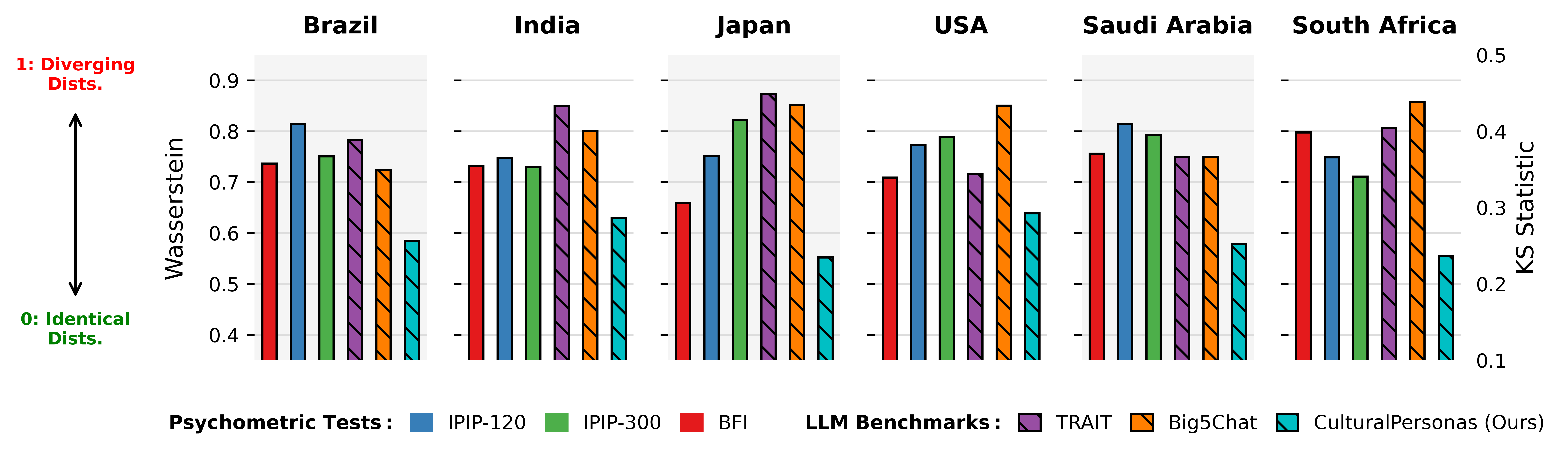}
        \label{fig:qwen_mcs_app}
    }

    \caption{Cross-cultural Wasserstein alignment in Multiple-Choice Selection (MCS) for Llama3 and Qwen2. Lower scores indicate closer alignment with human personality distributions. Overall, we observe \benchmark provides higher alignment across different countries.}
    \label{fig:app_mcs}
\end{figure*}

\begin{figure*}[t]
    \centering

    \subfigure[GPT-4o-mini]{
        \includegraphics[width=\textwidth]{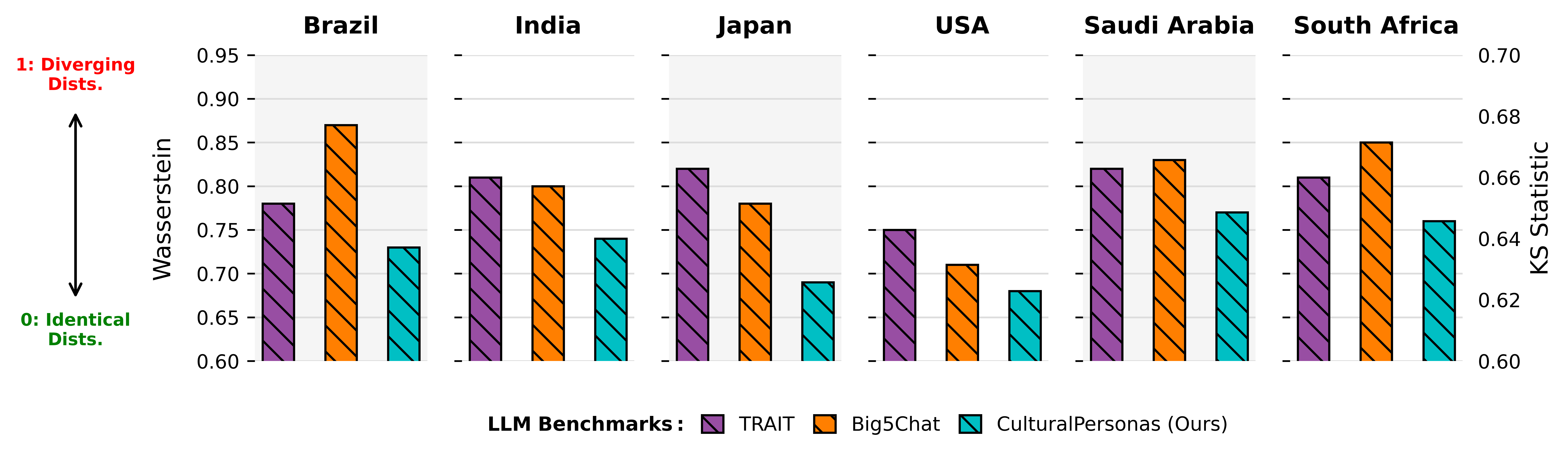}
        \label{fig:gpt_oeg_app}
    }

    \vspace{0.5em}

    \subfigure[Llama3]{
        \includegraphics[width=\textwidth]{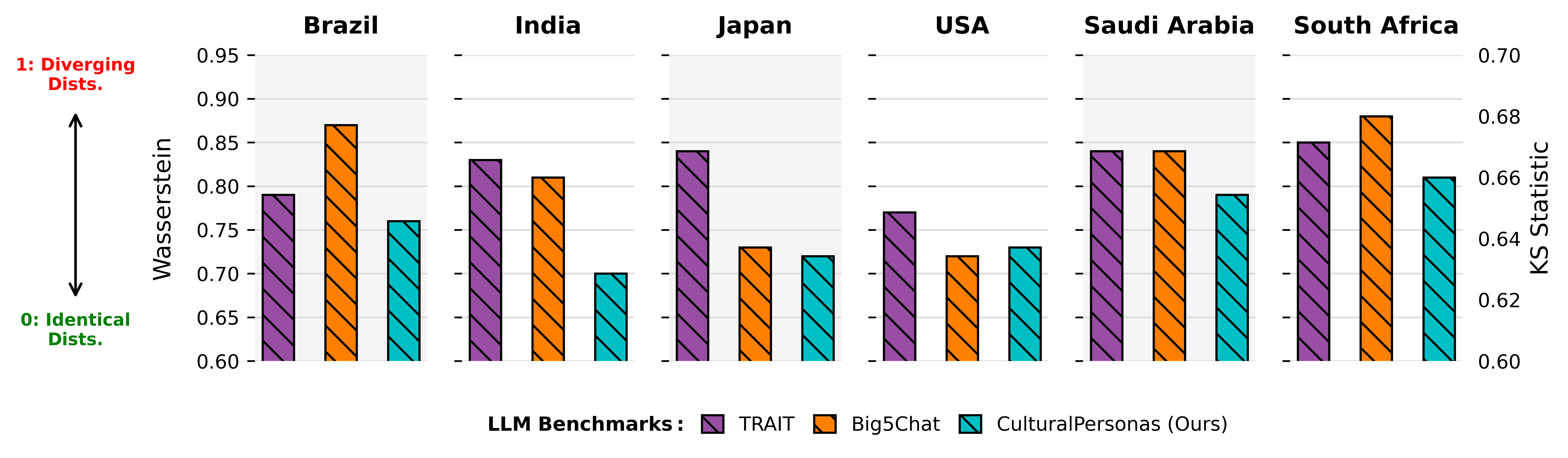}
        \label{fig:llama_oeg_app}
    }

    \vspace{0.5em}

    \subfigure[Qwen2]{
        \includegraphics[width=\textwidth]{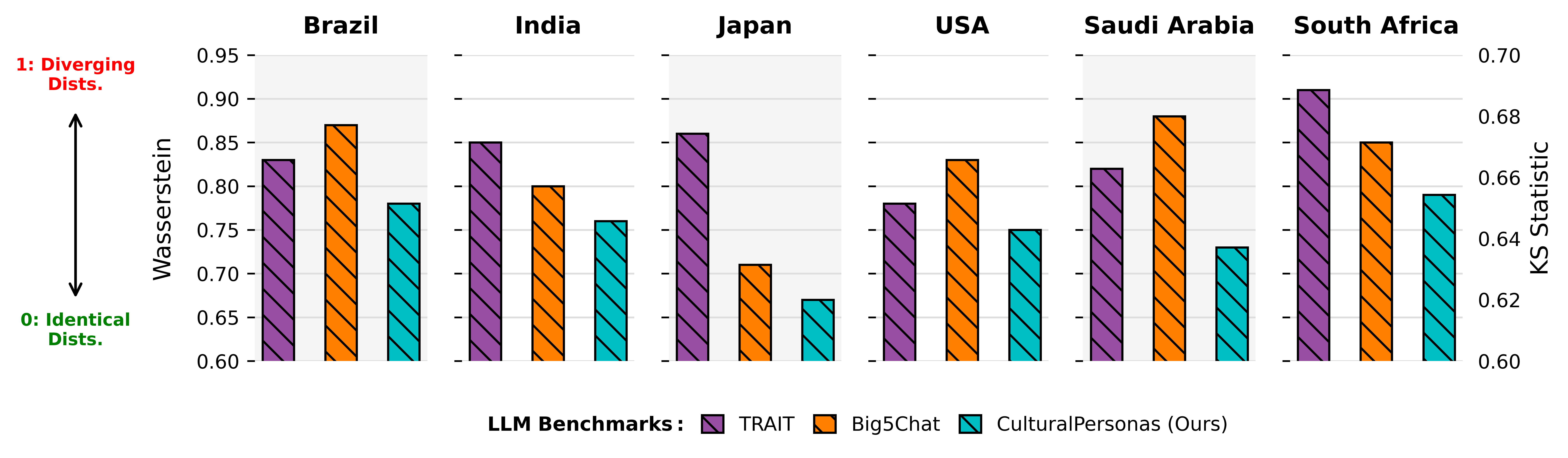}
        \label{fig:qwen_oeg_app}
    }

    \caption{Cross-cultural Wasserstein alignment in Open-Ended Generation (OEG) for all models. Lower scores indicate closer alignment with human personality distributions. We report higher scores in OEG than MCS, and models tend to diverge in performance much more. Although GPT-4o-mini yields best results overall, Llama3 tends to yield better results for India. Qwen2 tends to struggle with more diverse countries such as Brazil and South Africa.}
    \label{fig:app_oeg}
\end{figure*}

We first present all results for multiple choice selection (MCS) and open-ended generation (OEG) split by model in Figures \ref{fig:app_mcs} and \ref{fig:app_oeg}. Our results show that MCS setting is much easier with lower alignment metrics. However, in both settings, \benchmark can yield closer alignment with real human populations. GPT-4o-mini is able to achieve the strongest alignment across most countries. Qwen2 however tends to struggle more with expressive and underrepresented countries such as Brazil and South Africa. Llama3 tends to perform on average (in the middle) for most countries, only beating GPT-4o-mini on India. 

Next, we present the relative improvement rates for each country using \benchmark over both settings: MCS and OEG in Tables \ref{tab:app_mcs_rel} and \ref{tab:app_oeg_rel}. We find that the models generally show strong yields for \textit{Conscientiousness} and \textit{Openness}, with more moderate to low gains for traits such as \textit{Neuroticism} and \textit{Agreeableness}. In particular, gains are most pronounced in East Asian countries i.e. Japan, India, and Saudi Arabia and Africa i.e. South Africa. 

\begin{table*}[!t]
\centering
\scriptsize
\setlength{\tabcolsep}{2.5pt}
\begin{tabular}{@{}l | *{5}{r} | *{5}{r} | *{5}{r} | *{5}{r} | *{5}{r}@{}}
\toprule
\multirow{2}{*}{\textbf{Country}} & \multicolumn{5}{c|}{\textbf{BFI}} & \multicolumn{5}{c|}{\textbf{IPIP-120}} & \multicolumn{5}{c|}{\textbf{IPIP-300}} & \multicolumn{5}{c|}{\textbf{TRAIT}} & \multicolumn{5}{c}{\textbf{Big5Chat}} \\
 & O & C & E & A & N & O & C & E & A & N & O & C & E & A & N & O & C & E & A & N & O & C & E & A & N \\
\midrule
BR & 19.9 & 19.8 & 31.3 & 2.4 & 22.2 & 25.1 & 24.9 & 2.3 & 22.8 & 47.5 & 21.3 & 20.8 & 11.1 & 18.9 & 28.9 & 27.4 & 26.9 & 13.5 & 34.4 & 28.4 & 23.1 & 23.0 & 23.5 & 19.1 & 21.4 \\
IN & 16.0 & 15.7 & 7.5 & 34.6 & 0.2 & 17.4 & 16.5 & 22.0 & 6.4 & 17.7 & 16.0 & 15.2 & 3.0 & 32.2 & 6.7 & 27.6 & 27.4 & 20.0 & 23.2 & 35.8 & 24.8 & 24.4 & 31.0 & 19.6 & 18.7 \\
JP & 17.4 & 16.4 & 29.2 & 1.6 & 15.6 & 24.3 & 23.6 & 23.5 & 42.7 & 0.8 & 28.2 & 28.5 & 52.3 & 5.9 & 23.6 & 35.1 & 34.4 & 87.7 & 9.3 & 3.6 & 33.6 & 33.2 & 12.1 & 73.8 & 11.3 \\
US & 13.4 & 12.9 & 10.2 & 5.8 & 18.7 & 18.6 & 17.6 & 38.5 & 9.8 & 1.6 & 19.1 & 19.2 & 41.6 & 1.9 & 9.7 & 17.3 & 17.6 & 23.6 & 1.5 & 23.9 & 27.1 & 26.6 & 4.5 & 18.3 & 54.7 \\
SA & 22.5 & 21.9 & 30.7 & 14.3 & 16.2 & 25.7 & 25.4 & 5.8 & 31.4 & 36.8 & 24.5 & 24.3 & 31.8 & 27.0 & 10.5 & 25.6 & 25.2 & 47.5 & 22.4 & 0.8 & 25.1 & 25.2 & 11.4 & 1.9 & 58.4 \\
ZA & 26.7 & 26.3 & 28.0 & 8.4 & 40.0 & 23.1 & 23.3 & 0.6 & 58.8 & 7.2 & 20.9 & 20.3 & 5.2 & 52.3 & 0.4 & 31.4 & 30.4 & 26.5 & 3.1 & 58.7 & 34.1 & 33.3 & 17.6 & 56.7 & 22.9 \\

\bottomrule
\end{tabular}
\vspace{-1mm}
\caption{Relative improvement (\%) in Wasserstein distance using \benchmark across MCS settings for each \texttt{OCEAN} trait}
\label{tab:app_mcs_rel}
\end{table*}

\begin{table}[H]
\centering
\scriptsize
\setlength{\tabcolsep}{2.5pt}
\begin{tabular}{@{}l | *{5} {r} | *{5}{r} | *{5}{r} | *{5}{r} | *{5}{r}@{}}
\toprule
\multirow{2}{*}{\textbf{Country}} & \multicolumn{5}{c|}{\textbf{TRAIT}} & \multicolumn{5}{c|}{\textbf{Big5Chat}} \\
 & O & C & E & A & N & O & C & E & A & N \\
\midrule
BR & 6.4 & 7.2 & 3.4 & 4.8 & 5.2 & 16.4 & 17.5 & 10.8 & 11.3 & 9.0 \\
IN & 13.4 & 15.6 & 9.1 & 8.6 & 11.3 & 11.4 & 10.4 & 6.7 & 6.4 & 8.6 \\
JP & 20.4 & 17.5 & 14.5 & 14.2 & 20.4 & 6.8 & 7.2 & 6.9 & 4.7 & 5.4 \\ 
US & 8.8 & 7.9 & 3.4 & 2.3 & 5.0 & 7.0 & 6.6 & 3.0 & 2.5 & 4.2 \\
SA & 13.8 & 12.5 & 3.0 & 2.0 & 7.0 & 15.2 & 14.3 & 7.5 & 5.5 & 10.0 \\
ZA & 12.0 & 11.6 & 3.5 & 3.4 & 8.0 & 15.0 & 13.9 & 4.0 & 2.0 & 8.5 \\

\bottomrule
\end{tabular}
\vspace{-1mm}
\caption{Relative improvement (\%) in Wasserstein distance using \benchmark across OEG settings for each \texttt{OCEAN} trait}
\label{tab:app_oeg_rel}
\end{table}

\subsection{Analysis of Open-Ended Generation Results}
\label{sec:app_oeg}

In this section, we provide additional examples on various models express personality in the different benchmarked countries. In Table \ref{tab:gpt4o_ex}, we present examples on how GPT-4o-mini provides more culturally appropriate responses when prompted with \benchmark compared to other existing tools. 


\begin{table*}[t]
\centering
\footnotesize
\setlength{\tabcolsep}{4pt}
\begin{tabular}{p{1.75cm} p{2.2cm} p{9cm}} 
\toprule
\textbf{Country} & \textbf{Benchmark} & \textbf{Response} \\
\midrule

\multirow{3}{*}{Japan} 
  & TRAIT & I might profusely apologize to my colleague. \\
  & Big5Chat & I am very sorry. I did not intend to hurt you. \\
  & CulturalPersonas & I would bow deeply and express regret for hurting my colleague. \\

\midrule

\multirow{3}{*}{USA} 
  & TRAIT & I would say sorry and explain it was unintentional. \\
  & Big5Chat & I apologize and take responsibility to make it right. \\
  & CulturalPersonas & I would acknowledge my mistake and offer to fix it. \\

\midrule

\multirow{3}{*}{Brazil} 
  & TRAIT & I’d approach them and try to talk it out. \\
  & Big5Chat & Sorry! Let’s chat to smooth things over. \\
  & CulturalPersonas & I would show concern and invite a conversation. \\

\midrule

\multirow{3}{*}{India} 
  & TRAIT & I would offer to help with their tasks. \\
  & Big5Chat & I'm sorry. I didn’t mean to upset you. Can I make it up? \\
  & CulturalPersonas & I would acknowledge their hurt and apologize respectfully. \\

\midrule

\multirow{3}{*}{Saudi Arabia} 
  & TRAIT & I’d apologize and hope to maintain harmony. \\
  & Big5Chat & My apologies, I didn’t intend offense. \\
  & CulturalPersonas & I would sincerely apologize and show respect. \\

\midrule

\multirow{3}{*}{South Africa} 
  & TRAIT & I’d apologize and ask how to make things right. \\
  & Big5Chat & I'm sorry. Let me know how I can help. \\
  & CulturalPersonas & I would take responsibility and seek a positive way forward. \\

\bottomrule
\end{tabular}
\vspace{-1mm}
\caption{Illustrative examples of culturally appropriate personality expressions (GPT-4o-mini) across six countries for a workplace scenario testing \textit{Agreeableness}. \benchmark{} often provides culturally rich responses such as "bowing" in Japan, "respecting" in Saudi Arabia and India, and "showing concern" in Brazil.}
\label{tab:gpt4o_ex}
\end{table*}

\begin{table*}[!t]
\centering
\scriptsize
\begin{tabular}{p{5cm} p{5cm} p{5cm}}
\toprule
\multicolumn{3}{p{15cm}}{\textbf{Scenario:} You are at a friend's birthday party, and you only know a few people there. What do you do?} \\
\midrule
\textbf{Llama3} & \textbf{GPT-4o-mini} & \textbf{Qwen2} \\
\midrule
I’d not stick close to my friends, but I might feel shy even though I'd try to join into the groups.  
& I’d smile, start conversations with people nearby, and try to join in a group activity or dance. 
& I’d compliment someone’s outfit or joke about the music to break the ice, then gradually join the group. \\
\bottomrule
\end{tabular}
\vspace{-1mm}
\caption{Cultural expressions of \textit{Extraversion} in a Brazil-based social scenario. While all answers show moderate \textit{Extraversion}, Llama3 demonstrates a more indirect approach. GPT-4o-mini and Qwen2 align more with Brazilian sociability norms.}
\label{tab:extraversion_brazil}
\end{table*}

\begin{table*}[!t]
\centering
\scriptsize
\begin{tabular}{p{5cm} p{5cm} p{5cm}}
\toprule
\multicolumn{3}{p{15cm}}{\textbf{Scenario:} Your manager assigns extra work to you at the last minute, even though it wasn't your responsibility. How do you respond?} \\
\midrule
\textbf{Llama3} & \textbf{GPT-4o-mini} & \textbf{Qwen2} \\
\midrule
I would accept the task without complaint and do my best to complete it, showing commitment to team harmony. 
& I’d politely accept the work and later mention, with deference, that I’m happy to help this time but may struggle if this becomes frequent.
& I’d agree to take it on and say, “No worries, I’ve got it,” while gently suggesting we plan better next time to avoid last-minute stress. \\
\bottomrule
\end{tabular}
\vspace{-1mm}
\caption{Cultural expressions of \textit{Agreeableness} in a Japan-based workplace scenario. Llama3 and GPT-4o-mini reflect Japanese-style cooperation, while Qwen2 uses direct yet friendly feedback, aligned with US norms.}
\label{tab:agreeableness_japan}
\end{table*}

Additionally, we show examples of personality expression across various models. In Tables \ref{tab:extraversion_brazil} and \ref{tab:agreeableness_japan}, we show variations of how \textit{Extraversion} and \textit{Agreeableness} appear in Qwen2 v. Llama3 v. GPT-4o-mini. Qwen2 often struggles with expressing \textit{Agreeableness} in countries e.g. Japan. The example highlights more Western-centric norms in its answers compared to GPT-4o-mini and Llama3 whose responses highlight more of a passive and indirect method for maintaining harmony. On the other hand, Llama3 struggles with expressing \textit{Extraversion} in expressive nations such as Brazil. We can see when comparing with model responses from Qwen2 and GPT-4o-mini, these responses are much more expressive and engaging in ways related to Brazilian norms. 

\section{Additional Ablations}
\label{sec:app_ab}

In this section, we present ablation experiments by removing our selected persona prompts i.e. removing cultural priming and observing distributional alignment changes between models and human data \S\ \ref{sec:cultural_priming_ablations}. We also present results from using a simple prompt with only a country persona in \S\ \ref{sec:simple_prompts_ablations}.

\subsection{Cultural Priming Ablations}
\label{sec:cultural_priming_ablations}

\begin{figure*}[t]
    \centering

    \subfigure[Cultural Priming Ablations: MCS]{
        \includegraphics[width=\textwidth]{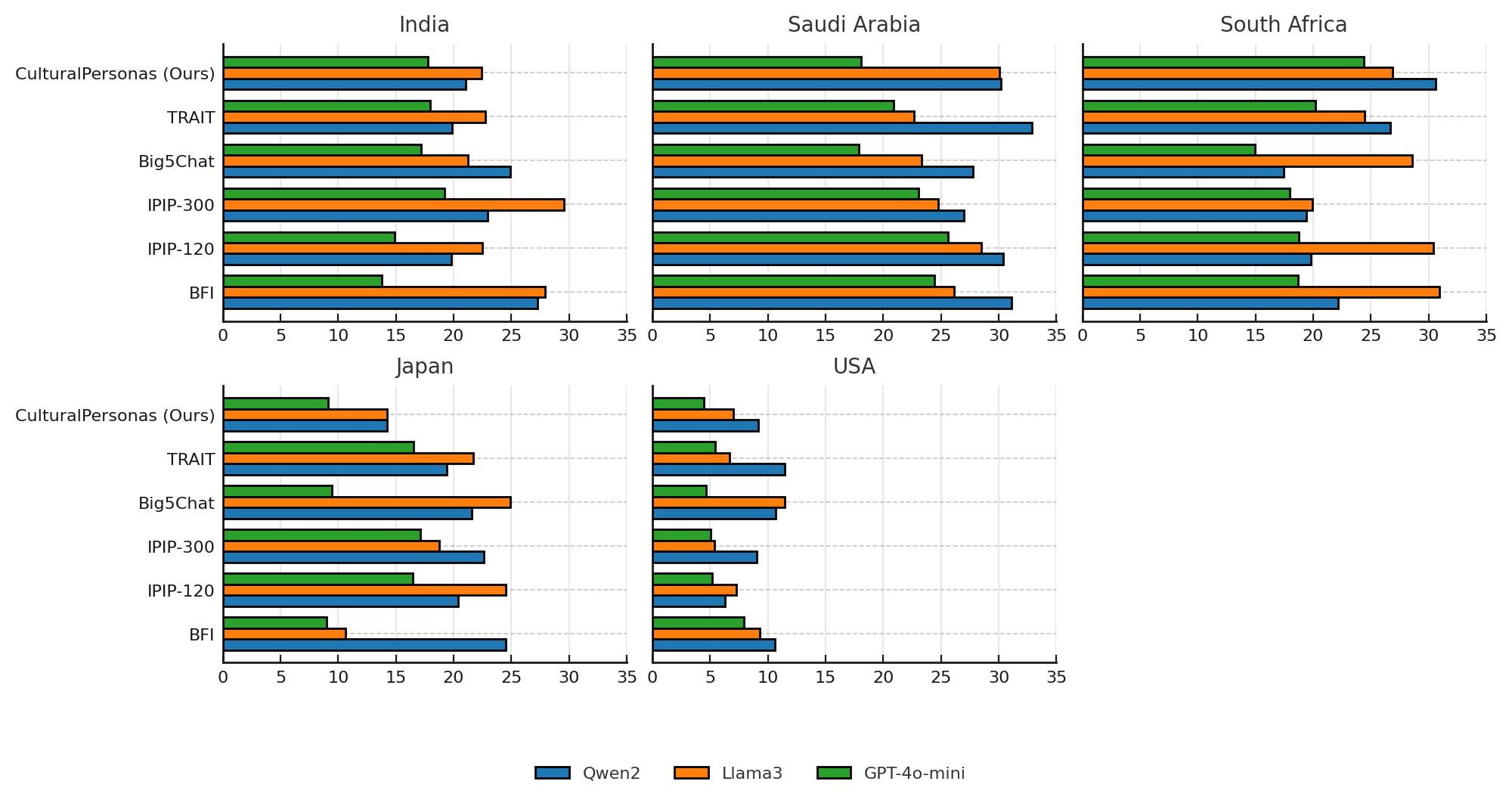}
        \label{fig:ablations_mcs_app}
    }

    \vspace{0.5em}

    \subfigure[Cultural Priming Ablations: OEG]{
        \includegraphics[width=\textwidth]{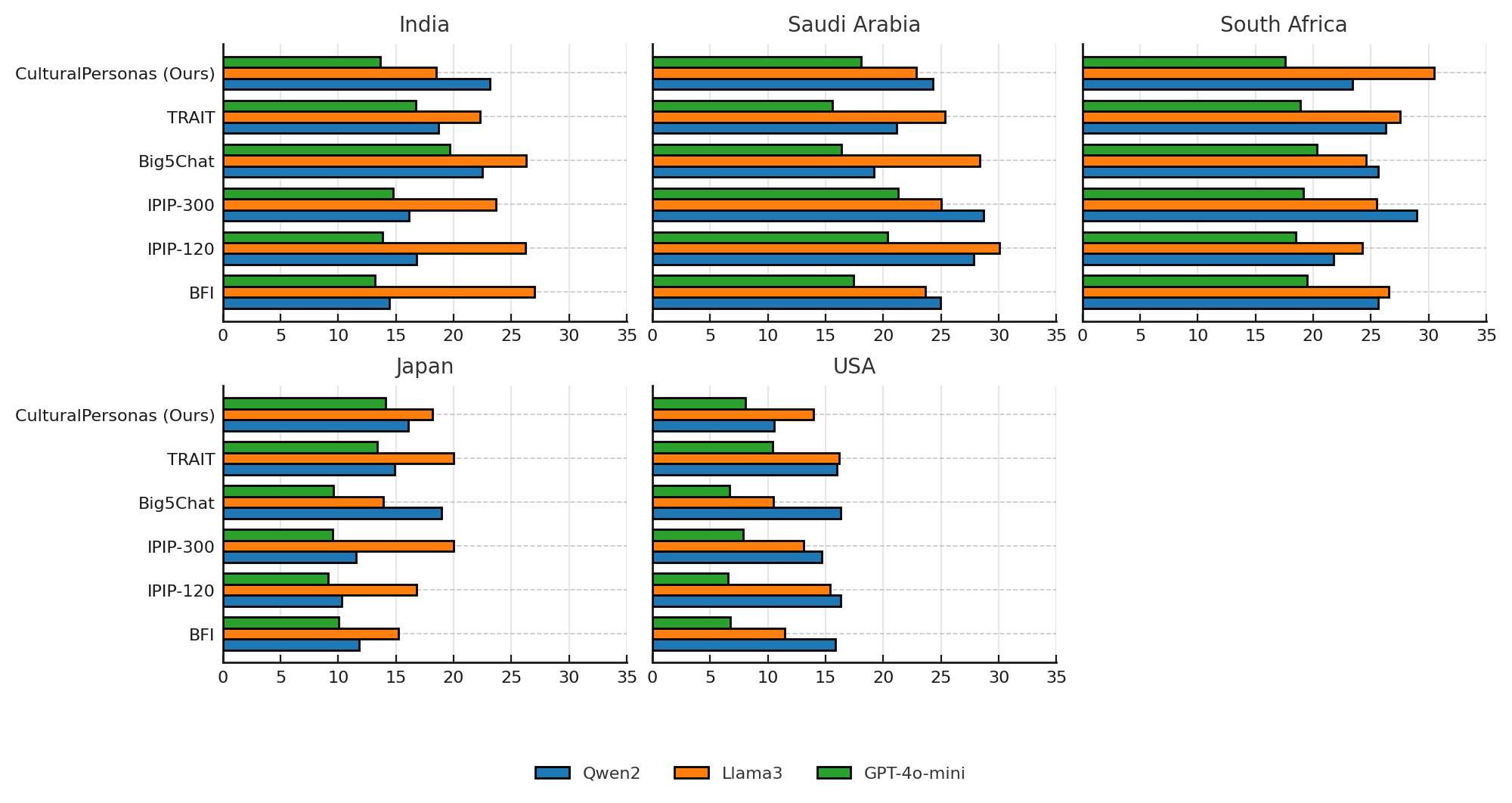}
        \label{fig:ablations_oeg_app}
    }

    \caption{Results from cultural priming ablations for each country and model for both MCS and OEG settings. We observe large increases (\%) in alignment (Wasserstein distances) for various countries including underrepresented and highly expressive nations.}
    \label{fig:app_ablations}
\end{figure*}

In this section, we provide results from all our cultural priming experiments (see Figure~\ref{fig:app_ablations}). We present deviations in performance (Wasserstein distances) for each country for each benchmarked model. We find that cultural priming indeed yields positive performance gains across all countries. Although priming for the USA provides the lowest gains, incorporating cultural norms still yields small, measurable improvements.

\subsection{Simple Prompting}
\label{sec:simple_prompts_ablations}

In this section, we provide results from our simple prompting experiments. Building on prior findings that simple persona prompts often lack sufficient context to elicit demographic-specific responses \cite{app1, app2}, we examine the extent to which cultural priming offers more effective guidance. In Table \ref{tab:simple_prompting_ablation}, we present the average Wasserstein distances for each country given a simple persona prompt i.e. \textit{``You are someone from \{country\}."}

\begin{table*}[!t]
\centering
\scriptsize
\setlength{\tabcolsep}{4pt}
\begin{tabular}{@{}l | r r r r r r@{}}
\toprule
\textbf{Country} & \textbf{BFI} & \textbf{IPIP-120} & \textbf{IPIP-300} & \textbf{TRAIT} & \textbf{Big5Chat} & \textbf{\benchmark}\\
\midrule
Brazil & 19.12 & 24.52 & 20.20 & 26.12 & 22.02 & 21.44 \\
India & 14.80 & 16.00 & 14.62 & 26.80 & 23.70 & 19.23 \\
Japan & 16.04 & 22.98 & 27.70 & 34.02 & 32.80 & 28.34 \\
USA & 12.20 & 17.22 & 18.30 & 16.78 & 26.24 & 31.34 \\
Saudi Arabia & 21.12 & 25.02 & 23.62 & 24.30 & 24.40 & 12.45 \\
South Africa & 25.88 & 22.60 & 19.82 & 30.02 & 32.92 & 20.23 \\
\bottomrule
\end{tabular}
\vspace{-1mm}
\caption{Average relative improvement (\%) in Wasserstein distance averaged across \texttt{OCEAN} traits using cultural priming over standard country context prompt for each country and evaluation set.}
\label{tab:simple_prompting_ablation}
\end{table*}

\end{document}